\documentclass[twoside,11pt]{article}
\usepackage{jair, theapa, rawfonts}

\jairheading{1}{1993}{1-15}{6/91}{9/91} 

\ShortHeadings{Topic-Aware Convolutional Neural Networks for Extreme
  Summarization} {Narayan, Cohen \& Lapata}
\firstpageno{1}

\usepackage{url}

\usepackage{amsmath}
\usepackage{mathtools}
\usepackage{amssymb}
\usepackage{multirow}

\usepackage{tree-dvips}
\usepackage{pgf,tikz}
\usepackage{tikz-qtree}
\usepackage{xcolor}
\usepackage{pgfplots}

\definecolor{forestgreen}{HTML}{009B55}
\definecolor{sepia}{HTML}{671800}
\definecolor{midnightblue}{HTML}{006795}
\definecolor{orangered}{HTML}{ED135A}

\newcommand{\newcite}[1]{\citeauthor{#1} \citeyear{#1}}
\newcommand{\citealt}[1]{\citeauthor{#1} \citeyear{#1}}
\newcommand{\citet}[1]{\citeauthor{#1} \citeyear{#1}}

\makeatletter
\newcommand{\thickhline}{%
    \noalign {\ifnum 0=`}\fi \hrule height 1pt
    \futurelet \reserved@a \@xhline
}
\makeatother

\begin{document}



\title{What is this Article about? Extreme Summarization with Topic-aware Convolutional Neural Networks}

\author{\name Shashi Narayan\thanks{The work was primarily done while Shashi was still at School of Informatics, University of Edinburgh.} \email shashinarayan@google.com \\
  \addr Google Research
  \AND
  \name Shay B. Cohen \email scohen@inf.ed.ac.uk \\
  \name Mirella Lapata \email mlap@inf.ed.ac.uk \\
  \addr Institute for Language, Cognition and Computation \\
  School of Informatics, University of Edinburgh}


\maketitle

\begin{abstract}
  \noindent We introduce \emph{extreme summarization}, a new
  single-document summarization task which aims at creating a short,
  one-sentence news summary answering the question ``What is the
  article about?''. We argue that extreme summarization, by nature, is
  not amenable to extractive strategies and requires an abstractive
  modeling approach. In the hope of driving research on this task
  further: (a)~we collect a real-world, large scale dataset by
  harvesting online articles from the British Broadcasting Corporation
  (BBC); and (b)~propose a novel abstractive model which is
  conditioned on the article's topics and based entirely on
  convolutional neural networks.  We demonstrate experimentally that
  this architecture captures long-range dependencies in a document and
  recognizes pertinent content, outperforming an oracle extractive
  system and state-of-the-art abstractive approaches when evaluated
  automatically and by humans on the extreme summarization
  dataset.\footnote{Our dataset, code, and demo are available at:
    \url{https://github.com/shashiongithub/XSum}.} 
  
\end{abstract}

\section{Introduction}

Automatic summarization is one of the central problems in Natural
Language Processing (NLP) posing several challenges relating to {\em
  understanding} (i.e.,~identifying important content) and {\em
  generation} (i.e.,~aggregating and rewording the identified content
into a summary). Of the many summarization paradigms that have been
identified over the years (see \citeR{mani2001automatic} and
\citeR{Nenkova:McKeown:2011} for comprehensive overviews)
single-document summarization has consistently garnered attention. 


\begin{figure}[t!]
  \center{\fontsize{10}{12}\selectfont 
    \begin{tabular}{ p{14.5cm} }
      \thickhline
      
      
      \textbf{\textsc{Summary:}} \emph{\textcolor{orangered}{A man and
          a child have been killed} after a \textcolor{sepia}{light}
        \textcolor{forestgreen}{aircraft made an emergency landing} on
        \textcolor{midnightblue}{a beach in Portugal}.}  
       \\ \hline
      
      \textbf{\textsc{Document:}} Authorities said the incident took place on \textcolor{midnightblue}{Sao Joao beach in Caparica, south-west of Lisbon}. \\ 

      The National Maritime Authority said \textcolor{orangered}{a middle-aged man and a young girl died} after they were unable to avoid the plane. \\
      
      
      The plane's only two occupants were unharmed, it added. \\

      The Diario de Noticias newspaper quoted an eyewitness who said the plane had been flying at a low altitude over the beach, although he did not realise anything was wrong until other beachgoers began running. \\

      One young witness told Reuters news agency: ``I was near the water when I saw the plane. I called my parents, the plane fell on the sand and ran over two people, fatally hurting them and another was injured, I think, but I'm not sure, people were running away.'' \\

      ``The plane is still there, but the ambulances arrived quickly. I think maybe the fuel ran out because I find it weird that it landed on the beach.'' \\

      Other reports said the victims had been sunbathing when \textcolor{forestgreen}{the plane made its emergency landing}. \\


      The Associated Press news agency said the girl who died had been with her parents, who were unhurt. The agency quoted witnesses from local television broadcasts. \\

      Joao Quadros, who was on the beach, tweeted photos of the aftermath, saying the plane had passed by his son by a matter of metres. There had been no noise, he said. \\ 

      Video footage from the scene carried by local broadcasters showed \textcolor{sepia}{a small recreational plane} parked on the sand, apparently intact and surrounded by beachgoers and emergency workers. One wing seemed to be misaligned in those photos. \\

      The cause of the emergency landing remains unclear. \\
        
      
      \thickhline

    \end{tabular}     
  }
  \caption{Example from our extreme summarization dataset
    showing a document and its one-line summary. Document content
    present in the summary is color-coded.
  }\label{fig:bbcex-1}
\end{figure}

Modern approaches to single document summarization are data-driven,
taking advantage of the success of neural network architectures and
their ability to learn continuous features without recourse to
preprocessing tools or linguistic annotations
\cite{jp-acl16,nallapati-signll16,see-acl17,Fan2017Controllable,paulus-socher-arxiv17,Pasunuru-multireward18,asli-multiagent18,narayan-rank18}. 
The application of neural networks to the summarization task has motivated
the development of large-scale datasets containing hundreds of
thousands of (news) document-summary pairs
\cite{nytcorpus,hermann-nips15,newsroom-naacl18}.  However, these
datasets often favor extractive models which create a summary by
identifying (and subsequently concatenating) the most important
sentences in a document \cite{jp-acl16,nallapati17,narayan-rank18}.
Abstractive approaches, despite being more faithful to the actual
summarization task --- professional editors employ various rewrite
operations to transform article sentences into a summary including
compression, aggregation, and paraphrasing \cite{Jing:2002} aside from
writing sentences from scratch --- they either lag behind extractive
ones or are mostly extractive, exhibiting a small degree of
abstraction
\cite{see-acl17,paulus-socher-arxiv17,Pasunuru-multireward18,asli-multiagent18,gehrmann-emnlp18}.


In this paper we introduce \textbf{extreme summarization}, a new
single-document summarization task which is not amenable to extractive
strategies and requires an abstractive modeling approach. The idea is
to create a short, one-sentence news summary answering the question
``What is this article about?''. Figure~\ref{fig:bbcex-1} shows an
example of a document and its extreme summary. As can be seen, the
summary is very different from a headline whose aim is to encourage
readers to read the story; it draws on information interspersed in
various parts of the document (not only the beginning) and displays
multiple levels of abstraction including paraphrasing, fusion,
synthesis, and inference. 

To drive research on abstractive summarization forward, we build a
dataset for the proposed task by harvesting online articles from the
British Broadcasting Corporation (BBC) that often include a
first-sentence summary. We further propose a novel deep learning model
which is well-suited to extreme summarization. Unlike most recent
abstractive approaches
\cite{rush-acl15,chenIjcai-16,nallapati-signll16,see-acl17,tanwan-acl17,paulus-socher-arxiv17,Pasunuru-multireward18,asli-multiagent18}
which rely on an encoder-decoder architecture modeled by recurrent
neural networks (RNNs), we present a \textbf{topic-conditioned} neural
model which is based entirely on \textbf{convolutional neural
  networks} \cite{convseq2seq}. Convolution layers capture long-range
dependencies between words in the document more effectively compared
to RNNs, allowing to perform document-level inference, abstraction,
and paraphrasing.  Our convolutional encoder associates each word with
a topic vector capturing whether it is representative of the
document's content, while our convolutional decoder conditions each
word prediction on a document topic vector capturing whether it is in
the theme of the document.

Experimental evaluation on the extreme summarization task shows that
our topic-aware convolutional model outperforms an oracle extractive
system (in terms of ROUGE) as well as state-of-the-art RNN-based
abstractive systems, a vanilla convolutional model \cite{convseq2seq}
and a convolutional model augmented with the pointer-generator
mechanism \cite{see-acl17}. We also conduct two human evaluations in
order to assess (a) which type of summary participants prefer and (b)
how much key information from the document is preserved in the
summary. Both evaluations overwhelmingly show that human subjects find
our summaries more informative and complete. To further illustrate
that the proposed model is generally applicable, we evaluate its
performance on the Newsroom Abstractive dataset
\cite{newsroom-naacl18}. Our experiments set a new state of the art
and highlight interesting differences between our extreme
summarization dataset and the Newsroom dataset.

Our contributions in this work are three-fold: we propose a new
single-document summarization dataset which encourages the development
of abstractive systems; we demonstrate through analysis and empirical
results that extractive approaches are not well-suited to the extreme
summarization task; and propose a novel topic-aware convolutional
sequence-to-sequence model for abstractive summarization. In the
remainder, we present an overview of related work
(Section~\ref{sec:related}) and the describe our extreme summarization
dataset in more detail
(Section~\ref{sec:bbcdataset}). Section~\ref{sec:topicconvabssum}
presents our model while Section~\ref{sec:results} discusses our
results.

\section{Related Work}
\label{sec:related}

\paragraph{Summarization Datasets}
The summarization of news articles has enjoyed wide popularity in
natural language processing due to its potential for various
information access applications which allow readers to spot emerging
trends, person mentions, the evolution of storylines, and so on. The
news domain has been the main focus of several Document Understanding
(DUC) and Text Analysis conferences (TAC) leading to the creation of
various high-quality summarization datasets
\cite{W04-1003,Over:2007:DUC}. More recently, the training
requirements of neural systems have led to the compilation of larger
datasets based on New York Times \cite{nytcorpus}, the Gigaword corpus
\cite{gigaword}, the CNN and DailyMail news outlets
\cite{hermann-nips15}, or a combination of several major news
publications \cite{newsroom-naacl18}. There has been some interest in
summarizing texts from other domains, such as longer scientific
articles \cite{Qazvinian2013GeneratingES,N18-2097,yasunaga.aaai19.scisumm}, Wikipedia articles
\cite{wiki:iclr:18}, live sport text commentary scripts
\cite{sportsnews:2016}, movie reviews
\cite{N16-1007,angelidis18summarizing} or online discussion forums and
blogs \cite{D15-1229,Kim:2018:arXiv,wikihow:2018}. In this paper, we
focus on generating extreme (single line abstractive) summaries for
BBC news articles.


The nature and quality of reference summaries vary for different
datasets. DUC datasets contain multi-reference summaries that are
manually written especially to evaluate summarization systems. Due to
the effort and cost involved in creating multiple reference summaries,
DUC datasets are rather small (few hundreds of articles) and fall
short of training neural summarization systems. Gigaword summaries are
short headlines \cite{rush-acl15}. Systems trained on New York Times
and CNN/DailyMail learn to generate multi-line abstracts or
highlights, however these summaries are mostly extractive and systems
trained on them unavoidably learn to perform mainly copying operations
even when capable of performing abstraction
\cite{see-acl17,newsroom-naacl18}. Newsroom \cite{newsroom-naacl18}
summaries are manual descriptions of news articles writen by authors
and editors in newsrooms of 38~major news publications. Coming from a
variety of sources, these summaries exhibit different degrees of
abstraction, they are not visible to readers but are often used to
index the article \cite{newsroom-naacl18}. In contrast, our summaries
are read together with the article, they are the first sentence
readers see (often highlighted in boldface) prior to digesting the
full article.  The Newsroom dataset is fairly large containing~1.3 million
articles and their summaries and goes some way towards addressing the
concerns relating to biases towards extractive strategies in earlier
datasets. We discuss the differences between Newsroom and our dataset
in more detail in Section~\ref{sec:bbcdataset} and also present
experimental results with our model in
Section~\ref{sec:results-newsroom}.

\paragraph{Summarization Approaches}

Approaches to document summarization fall under two major paradigms:
extractive systems select sentences from the document and assemble
them together to generate a summary, while abstractive systems create
a summary from scratch, possibly generating new words or phrases which
are not in the document.

A great deal of previous work has focused on extractive summarization
which is usually modeled as a sentence ranking or binary
classification problem (i.e., sentences which are top ranked or
predicted as \texttt{True} are selected as summaries).  Early attempts
mostly leverage human-engineered features including sentence position
and length \shortcite{radev-lrec2004}, keywords and the presence of
proper nouns
\cite{Kupiec:1995binary,mani2001automatic,SparckJones:2007},
information based on frequency \cite{nenkova-06} or events
\cite{filatova-04event}. These methods often learn to score each
sentence independently
\cite{Barzilay97usinglexical,Teufel97sentenceextraction,Barzilay:2002:ISS,erkan:2004:lexrank,Mihalcea04TextRank,Shen:2007:IJCAI,Schilder:2008:fastsum,Wan:2010:urank},
however summary quality can be improved heuristically
\cite{Yih:2007:MSM,Dorr03}, via max-margin methods
\cite{Carbonell:1998:UMD,Li:2009:EDC}, or integer-linear programming
\cite{woodsend-acl10,berg:2011,Woodsend:2012:emnlp,Almeida:acl13,parveen:2015:tgraph,martins-smith:2009:ILPNLP}.

Modern extractive summarization models
\cite{krageback-cvsc14,Yin-ijcai15,jp-acl16,nallapati17} are
data-driven and learn continuous features using neural network
architectures without any linguistic preprocessing or reliance on
expert feature design. The majority of them conceptualize extractive
summarization as a sequence labeling task in which each label
specifies whether each document sentence should be included in the
summary
\cite{jp-acl16,nallapati17,narayan-arxiv17,Yasunaga:2017:gcn,narayan-sidenet18,narayan-rank18,zhang-EtAl:2018:EMNLP3,mendes-etal-2019-jointly}.
These models often rely on recurrent neural networks to derive a
meaning representation of the document which is then used to label
each sentence, taking the previously labeled sentences into account.

There has also been a surge of interest in neural network models for
abstractive summarization which is viewed as a sequence-to-sequence
problem \cite{sutskever-nips14,cho-emnlp14,bahdanau-arxiv14}. Central
in most approaches
\cite{rush-acl15,chenIjcai-16,nallapati-signll16,tanwan-acl17,see-acl17}
is an encoder-decoder architecture modeled by recurrent neural
networks. The encoder reads the source sequence into a list of
continuous-space representations from which the decoder generates the
target sequence. \citeA{see-acl17} refine this sequence-to-sequence
architecture with a copy mechanism \cite{vinyals-nips15,gu-EtAl:2016}
which allows to reuse sequences from the source document and with a
coverage mechanism \cite{tu-EtAl:2016:P16-1} which allows to keep
track of what has been summarized, discouraging repetition.  A few
extractive \cite{narayan-rank18,wuhu-aaai18,dong-EtAl:2018:EMNLP} and
abstractive
\cite{paulus-socher-arxiv17,li-arxiv-18,asli-multiagent18,Pasunuru-multireward18,chen-bansal:2018:Long,kryciski-EtAl:2018:EMNLP}
approaches obtain performance improvements by combining the
maximum-likelihood cross-entropy loss with rewards from policy
gradient reinforcement learning \cite{Williams:1992} to directly
optimize the evaluation metric relevant for the summarization task.
 
Our topic-conditioned convolutional model differs from earlier
approaches both in application and formulation.  Unlike abstractive
models based on recurrent neural networks, we adopt a fully
convolutional endcoder-decoder architecture \cite{convseq2seq}.
Convolution layers capture long range dependencies between words in
the document more effectively compared to RNNs, allowing to perform
document-level inference, abstraction, and paraphrasing. Our
convolutional encoder associates each word with a topic vector
capturing whether it is representative of the document’s content,
while our convolutional decoder conditions each word prediction on a
document topic vector. Convolutional alternatives to sequence modeling
have been proposed for machine translation \cite{convenc_mt}, headline
generation \cite{convseq2seq}, and story generation
\cite{fan-hier-gen}, however we are not aware of any prior work
targetting summarization. The Transformer architecture
\cite{transformer} presents an alternative to convolutions, also
aiming at eliminating the fundamental constraint of sequential
computation, and has been successfully applied to sentence and document
summarization \cite{wiki:iclr:18,liu_arxiv19,mass_icml19,unilm_arxiv19}.\footnote{Experiments with Transformer architectures are outside the scope of this paper. Recent work \cite{laura_aclshort19} on  
multiodocument summarization shows that Transformer-based models  
perform on par with their convolutional alternatives.}

Our convolutional model uses topic vectors to foreground salient words
in the document. The idea is inspired from traditional summarization
methods for {\em content selection}
\cite{mani2001automatic,radev-lrec2004,nenkova-06,SparckJones:2007},
however, our topics are not manually crafted, they are automatically
learned using an LDA model \cite{Blei:2003:LDA}. Several recent summarization
models have explored  architectures dedicated to content
selection; \newcite{li-EtAl:2018:N18-2} extract a set of keywords from
the document to guide the summarization process. \newcite{P17-1101}
and \newcite{li-EtAl:2018:EMNLP2} use dedicated gates to filter the
representation of the source document; while others modulate the
attention based on how likely it is for a word or a sentence to be
included in a summary
\cite{N18-2097,hsu-EtAl:2018:Long,gehrmann-emnlp18} or use
reinforcement learning to optimize content selection objectives
\cite{Pasunuru-multireward18,chen-bansal:2018:Long}. Document-level
semantic information (as expressed via latent topics) has been
previously integrated with recurrent neural networks
\cite{mikolovZ12,Ghosh2016ContextualL,dieng-iclr17}, however, we are
not aware of any existing convolutional models.

\section{The XSum Dataset}
\label{sec:bbcdataset}

In this section we present, XSum, our extreme summarization dataset
which consists of BBC articles and accompanying single sentence
summaries. We describe how XSum was obtained, provide comparisons with
popular summarization benchmarks, and analyze how it differs from them
both quantitatively and qualitatively.

\subsection{Data Collection} 
Each BBC article is prefaced with an introductory sentence (aka summary)
which is professionally written, typically by the author of the
article. The summary bears the HTML class
``story-body\_\_introduction,'' and can be easily identified and
extracted from the main text body (see Figure~\ref{fig:bbcex-1} for an
example summary-article pair).


To create a large-scale dataset for extreme summarization, we followed
the methodology proposed in \newcite{hermann-nips15}. Specifically, we
collected~226,711 Wayback archived BBC articles ranging over almost a
decade (2010 to 2017) and covering a wide variety of domains
(e.g.,~News, Politics, Sports, Weather, Business, Technology, Science,
Health, Family, Education, Entertainment, and Arts). Each article comes
with a unique identifier in its URL, which we used to randomly split
the dataset into training (90\%, 204,045), validation (5\%, 11,332),
and test (5\%, 11,334) set.

\begin{table*}[t!]
  \begin{center}{
  \begin{tabular}{l | r r r}
    \thickhline
    \multirow{2}{*}{Datasets} & \multicolumn{3}{c}{Corpus Size (\# docs)}\\
    & training & validation & test \\ \thickhline 
    CNN & 90,266 & 1,220 & 1,093 \\
    DailyMail & 196,961 & 12,148 & 10,397  \\ 
    NY Times & 589,284 & 32,736 & 32,739 \\ 
    Newsroom & 992,966 & 108,591 & 108,650 \\ 
    Newsroom-Mixed & 328,634 & 35,879 & 36,006  \\
    Newsroom-Ext & 331,778 & 36,332 & 36,122  \\
    Newsroom-Abs & 332,554 & 36,380 & 36,522  \\ 
    XSum & 204,045 & 11,332 & 11,334  \\ \thickhline
  \end{tabular}}
  \end{center}
  \caption{Comparison of XSum with benchmark summarization datasets: CNN and
    DailyMail datasets \protect\cite{hermann-nips15}, NY Times 
    \protect\cite{nytcorpus}, and Newsroom \protect\cite{newsroom-naacl18}.
    We present the full Newsroom
    dataset (Newsroom) and its three subsets: mostly extractive
    (Newsroom-Ext), mostly abstractive
    (Newsroom-Abs), and mixed (Newsroom-Mixed). We report  corpus
    size, i.e., the number of documents in,
    training, validation, and test sets.
  } \label{table:bbc-size-comparison-1}
\end{table*}

\begin{table*}[t!]
  \begin{center}{
  \begin{tabular}{ l | c c | c c | r r}
    \thickhline
    \multirow{2}{*}{Datasets} & \multicolumn{2}{c|}{avg. document length} & \multicolumn{2}{c|}{avg. summary length} & \multicolumn{2}{c}{vocabulary size}\\
    & words & sentences & words & sentences & document & summary\\ \thickhline 
    CNN & 760.50 & 33.98 & 45.70 & 3.59 & 343,516 & 89,051 \\
    DailyMail & 653.33 & 29.33 & 54.65 & 3.86 & 563,663 & 179,966 \\ 
    NY Times & 800.04 & 35.55 & 45.54 & 2.44 & 1,399,358 & 294,011 \\ 
    Newsroom & 770.09 & 34.73 & 30.36 & 1.43 & 2,646,681 & 360,290\\ 
    Newsroom-Mixed & 830.58 & 36.63 & 23.78 & 1.17 & 1,271,435 & 169,875 \\
    Newsroom-Ext & 706.06 & 31.65 & 45.78 & 1.88 & 1,214,748 & 243,062 \\
    Newsroom-Abs & 774.17 & 35.92 & 21.49 & 1.25 & 1,385,205 & 157,939 \\ 
    XSum & 431.07 & 19.77 & 23.26 & 1.00 & 399,147 & 81,092 \\ \thickhline
  \end{tabular}}
  \end{center}
  \caption{We compare datasets with respect to average document
    (source) and summary (target) length (in terms of words and
    sentences), and vocabulary size on both on source and target. See
    main text
    for steps
    taken to split and pre-process these datasets.  For the
    vocabulary, we lower case
    tokens.} \label{table:bbc-size-comparison-2}
\end{table*}

Tables~\ref{table:bbc-size-comparison-1}
and~\ref{table:bbc-size-comparison-2} compare XSum with the CNN,
DailyMail, NY Times, and Newsroom benchmarks. For CNN and DailyMail,
we used the original splits of \protect\newcite{hermann-nips15} and
followed \protect\newcite{narayan-rank18} to preprocess them. For NY
Times \protect\cite{nytcorpus}, we used the splits and pre-processing
steps of \protect\newcite{paulus-socher-arxiv17}. For the Newsroom
dataset, we used the splits and pre-processing steps of
\protect\newcite{newsroom-naacl18}.  We present comparisons with the
full Newsroom dataset (Newsroom) and its three subsets: mostly
extractive (Newsroom-Ext), mostly abstractive (Newsroom-Abs), and
mixed (Newsroom-Mixed). As can be seen in
Table~\ref{table:bbc-size-comparison-1}, XSum contains a substantial
number of training instances, similar to DailyMail; documents and
summaries in XSum are shorter in relation to most datasets (see
Table~\ref{table:bbc-size-comparison-2}) but the vocabulary size is
sufficiently large, comparable to CNN.

\subsection{How Abstractive is XSum?} 


To support the claim that XSum summaries are fairly abstractive and as
a result systems trained on them could not resort to extractive
strategies, we record the percentage of novel $n$-grams in the gold
summaries that do not appear in their source documents. As shown in
Table~\ref{table:ngram-coverage-lead-oracle-1}, there are 36\% novel
unigrams in the XSum reference summaries compared to 17\% in CNN,
17\%~in DailyMail, 23\%~in NY Times, and 18\%~in Newsroom.  This
indicates that XSum summaries are more abstractive. The proportion of
novel constructions grows for larger $n$-grams across datasets,
however, it is much steeper in XSum whose summaries exhibit
approximately 83\% novel bigrams, 96\% novel trigrams, and 98\% novel
4-grams (comparison datasets display around 47--55\%~new bigrams,
58--72\%~new trigrams, and \mbox{63--80}\%~novel 4-grams).

\begin{table*}[t!]
  \begin{center}{
  \begin{tabular}{ l | c c c c } 
    \thickhline
    \multirow{2}{*}{Datasets} & \multicolumn{4}{c}{\% of novel n-grams in gold summary} \\
    & unigrams & bigrams & trigrams & 4-grams \\ \thickhline 
    CNN & 16.75 & 54.33 & 72.42 & 80.37 \\
    DailyMail & 17.03 & 53.78 & 72.14 & 80.28 \\ 
    NY Times & 22.64 & 55.59 & 71.93 & 80.16 \\ 
    Newsroom & 18.31 & 46.80 & 58.06 & 62.72  \\ 
    Newsroom-Mixed & 13.78 & 48.37 & 67.15 & 77.11 \\ 
    Newsroom-Ext & 2.65 & 7.25 & 10.25 & 12.42  \\
    Newsroom-Abs & {38.25} & {84.36} & {96.39} & 98.28 \\ 
    XSum & 35.76 & 83.45  & 95.50  & {98.49}  \\ \thickhline
  \end{tabular}}
  \end{center}
  \caption{Proportion of novel $n$-grams in gold summaries for CNN,
    DailyMail, NY Times, Newsroom, and XSum datasets.  All results
    are computed on the test set.} \label{table:ngram-coverage-lead-oracle-1}
\end{table*}

We further evaluate two extractive methods, \textsc{lead} and
\textsc{ext-oracle}, on these
datasets. 
\textsc{lead} is often used as a strong lower bound for news
summarization \cite{nenkova-05} and creates a summary by selecting the
first few sentences or words in the document. We extracted the first
3~sentences for CNN documents and the first 4~sentences for DailyMail
\cite{narayan-rank18}. Following previous work
\cite{durrett-nyt-ext,paulus-socher-arxiv17}, we obtained
\textsc{lead} summaries based on the first 100 words for NY Times
documents. For Newsroom, we extracted the first 2~sentences to form
the \textsc{lead} summaries.  For XSum, we selected the first sentence
in the document (excluding the one-line summary) to generate the
\textsc{lead}. Our second method, \textsc{ext-oracle}, can be viewed
as an upper bound for extractive models
\cite{nallapati17,narayan-rank18}. It creates an oracle summary by
selecting the best possible set of sentences in the document that
gives the highest ROUGE \cite{rouge} with respect to the gold
summary. For XSum, we simply selected the single-best sentence in the
document as summary.

\begin{table*}[t!] 
  \begin{center}{ 
  \begin{tabular}{ l | c c c | c c c } 
    \thickhline
    \multirow{2}{*}{Datasets} & \multicolumn{3}{c|}{\textsc{lead}} & \multicolumn{3}{c}{\textsc{ext-oracle}} \\
    & R1 & R2 & RL & R1 & R2 & RL \\ \thickhline 
    CNN & 29.15 & 11.13 & 25.95 & 50.38 & 28.55 & 46.58 \\ 
    DailyMail & 40.68 & 18.36 & 37.25 & 55.12 & 30.55 & 51.24 \\ 
    NY Times & 31.85 & 15.86 & 23.75 & 52.08 & 31.59 & 46.72 \\ 
    Newsroom & 33.04 & 22.35 & 30.31 & 57.09 & 42.94 & 53.65 \\
    Newsroom-Mixed & 27.95 & 13.87 & 23.97 & 51.98 & 34.04 & 46.96 \\ 
    Newsroom-Ext & 55.87 & 50.60 & 54.76 & 89.63 & 87.20 & 89.32 \\
    Newsroom-Abs & {15.44} & \hspace*{1ex}2.72 & 12.32 & 29.85 & \hspace*{1ex}{7.82} & 24.86\\ 
    XSum & 16.30 & \hspace*{1ex}{1.61}  & {11.95}  & {29.79}  & \hspace*{1ex}8.81  & {22.65}  \\ \thickhline
  \end{tabular}}
  \end{center}
  \caption{Performance of extractive baselines on CNN,
    DailyMail, NY Times, Newsroom, and XSum datasets. We report ROUGE
    scores for the \textsc{lead} baseline and \textsc{ext-oracle}, the
    extractive oracle system.  All results are computed on the test
    set.} \label{table:ngram-coverage-lead-oracle-2}
\end{table*}

Table~\ref{table:ngram-coverage-lead-oracle-2} reports the performance
of the two extractive methods using ROUGE-1 (R1), ROUGE-2 (R2), and
ROUGE-L (RL) with the gold summaries as reference. The \textsc{lead}
baseline performs extremely well on CNN, DailyMail, NY Times and
Newsroom confirming that they contain fairly extractive summaries.
\textsc{ext-oracle} further shows that improved sentence selection
would bring further performance gains to extractive
approaches. Abstractive systems trained on these datasets often have a
hard time beating the \textsc{lead}, let alone \textsc{ext-oracle}, or
display a low degree of novelty in their summaries
\cite{see-acl17,tanwan-acl17,paulus-socher-arxiv17,Pasunuru-multireward18,asli-multiagent18,gehrmann-emnlp18}.
Interestingly, \textsc{lead} and \textsc{ext-oracle} perform poorly on
XSum underlying the fact that it contains genuinely abstractive
summaries. 




\newcite{newsroom-naacl18} also find that CNN / Daily Mail and New
York Times are skewed towards extractive summaries (albeit following
different analysis metrics). The abstractive subset of their Newsroom
dataset (Newsroom-Abs) demonstrates similar patterns to XSum in terms
of the percentage of novel $n$-grams in the gold summary and the
performance of extractive methods (\textsc{lead} and
\textsc{ext-oracle}). However, XSum differs from Newsroom in two key
respects. Firstly, Newsroom is a fairly diverse dataset, it contains
documents and summaries from multiple news outlets representing a
large range of summarization styles from highly abstractive to highly
extractive, while XSum is not; it covers a single news outlet
(i.e.,~BBC) and a uniform summarization style (i.e.,~a single
sentence). Another difference comes from the way the reference
summaries are extracted in these two datasets. Newsroom summaries are
extracted using the HTML meta-tag ``description,'' and constitute
descriptions of the document's content which are often used for
indexing but are not shown to the readers. In comparison, XSum
summaries are aimed at the reader and meant to be read together with
the article. Newsroom summaries are often indicative -- they provide
merely an indication of the subject matter of the document without
giving away detail on its content. In contrast, XSum summaries are
more informative, they contain pertinent information necessary to
convey the gist of the document.  We further explore these differences
in our experimental evaluation (see
Section~\ref{subsec:contentindicative}).

\section{Topic-Aware Convolutional Model for Summarization}
\label{sec:topicconvabssum}

Unlike tasks like machine translation and paraphrase generation where
there is often a one-to-one semantic correspondence between source and
target words, document summarization must distill the content of a
document into a few important facts. This is even more challenging for
our task, where the compression ratio is extremely high, and pertinent
content can be easily missed.

Our model builds on the work of \newcite{convseq2seq} who develop an
encoder-decoder architecture with an attention mechanism
\cite{NIPS2015_5846} based exclusively on deep convolutional networks.
Their convolutional alternative to sequence modeling has shown promise
for machine translation \cite{convenc_mt,convseq2seq} and story
generation \cite{fan-hier-gen}. We believe that convolutional
architectures are attractive for extreme summarization for at least
two reasons. Firstly, contrary to recurrent networks which view the
input as a chain structure, convolutional networks can be stacked to
represent large context sizes. Secondly, hierarchical features can be
extracted over larger and larger contents, allowing to represent
long-range dependencies efficiently through shorter paths.

We adapt this model to our task by allowing it to recognize pertinent
content (i.e.,~by foregrounding salient words in the document). In
particular, we improve the convolutional encoder by associating each
word with a vector representing topic salience, and the convolutional
decoder by conditioning each word prediction on the document topic
vector. Our model aims to generate informative summaries that are
grounded in the input document and its content.



\begin{figure}[th!]
  \center{
    \begin{tikzpicture}[scale=0.62]
      
      \begin{scope}[shift={(0,0)}] 
        \draw [rounded corners=5pt] (-2.5,1.75) rectangle (15,-21.5);
        
        \draw [fill=sepia,opacity=0.1] (0,0) rectangle (1.5,-0.5);
        \draw [fill=sepia,opacity=0.3] (1.5,0) rectangle (6,-0.5);
        \draw [fill=sepia,opacity=0.1] (6,0) rectangle (7.5,-0.5);
        \draw [fill=orangered,opacity=0.1] (0,-0.5) rectangle (1.5,-1);
        \draw [fill=orangered,opacity=0.3] (1.5,-0.5) rectangle (6,-1);
        \draw [fill=orangered,opacity=0.1] (6,-0.5) rectangle (7.5,-1);
        
        \draw [gray] (0,-0.5) -- (7.5,-0.5);
        \draw [gray] (1.5,0) -- (1.5,-1);
        \draw [gray] (3,0) -- (3,-1);
        \draw [gray] (4.5,0) -- (4.5,-1);
        \draw [gray] (6,0) -- (6,-1);
        \node at (8.5,-0.75) {$x_i+p_i$};
        \node at (8.5,-0.25) {$t'_i \otimes t_D$};  
        \node at (-0.5,-0.5) {$\mathbf{e}$};

        \node at (1,0.75) {\small \rotatebox{45}{[PAD]}}; 
        \node at (2.5,0.75) {\small \rotatebox{45}{England}}; 
        \node at (4,0.75) {\small \rotatebox{45}{won}}; 
        \node at (5.25,0.75) {\small \rotatebox{45}{.}}; 
        \node at (7,0.75) {\small \rotatebox{45}{[PAD]}}; 
        
        
        \node at (0,-2.5) {Convolutions};
        
        \draw [fill=midnightblue,opacity=0.2] (3.75,-1.5) -- (6.75, -1.5) -- (5.25, -3.5) -- (3.75,-1.5);
        \draw [fill=midnightblue,opacity=0.2] (2.25,-1.5) -- (5.25, -1.5) -- (3.75, -3.5) -- (2.25,-1.5);
        \draw [fill=midnightblue,opacity=0.2] (0.75,-1.5) -- (3.75, -1.5) -- (2.25, -3.5) -- (0.75,-1.5);
        
        \draw [fill=white,opacity=1] (1.75,-3.25) rectangle (2.75,-3.75);
        \draw [fill=white,opacity=1] (3.25,-3.25) rectangle (4.25,-3.75);
        \draw [fill=white,opacity=1] (4.75,-3.25) rectangle (5.75,-3.75);
        
        \draw (2.25,-3.25) -- (2.25,-3.75);
        \draw (3.75,-3.25) -- (3.75,-3.75);
        \draw (5.25,-3.25) -- (5.25,-3.75);
        

        \node at (0,-4.45) {GLU};
        
        \draw [->] (2,-3.75) -- (2,-4.25);
        \draw [gray] (2,-4.45) circle (0.2cm);
        \node at (2, -4.45) {$f$};
        \draw [->] (2.5,-3.75) -- (2.5,-5);
        \node at (2.5, -5.2) {$\otimes$};
        \draw [->] (2, -4.65) -- (2, -5.2) -- (2.25,-5.2);
        \draw [->] (2.5,-5.4) -- (2.5,-6);

        \draw [->] (3.5,-3.75) -- (3.5,-4.25);
        \draw [gray] (3.5, -4.45) circle (0.2cm);
        \node at (3.5, -4.45) {$f$};
        \draw [->] (4,-3.75) -- (4,-5);
        \node at (4,-5.2) {$\otimes$};
        \draw [->] (3.5,-4.65) -- (3.5,-5.2) -- (3.75,-5.2);
        \draw [->] (4,-5.4) -- (4,-6);

        \draw [->] (5,-3.75) -- (5,-4.25);
        \draw [gray] (5, -4.45) circle (0.2cm);
        \node at (5, -4.45) {$f$};
        \draw [->] (5.5,-3.75) -- (5.5,-5);
        \node at (5.5, -5.2) {$\otimes$};
        \draw [->] (5, -4.65) -- (5,-5.2) -- (5.25,-5.2);
        \draw [->] (5.5,-5.4) -- (5.5,-6);

        \draw [fill=midnightblue,opacity=0.2] (1.75,-6) rectangle (3.25,-7);
        \draw [fill=midnightblue,opacity=0.2] (3.25,-6) rectangle (4.75,-7);
        \draw [fill=midnightblue,opacity=0.2] (4.75,-6) rectangle (6.25,-7);
        \node at (1.25,-6.5) {$\mathbf{z^u}$};

        \node at (-1,-8.75) {\rotatebox{90}{Attention}}; 

        \draw [->] (2.5,-7) -- (2.5,-8);
        \draw [->] (4,-7) -- (4,-8);
        \draw [->] (5.5,-7) -- (5.5,-8);

        \draw [fill=forestgreen,opacity=0.2] (1.75,-8) rectangle (3.25,-8.5);
        \draw [fill=forestgreen,opacity=0.8] (3.25,-8) rectangle (4.75,-8.5);
        \draw [fill=forestgreen,opacity=0.7] (4.75,-8) rectangle (6.25,-8.5);
        \draw [fill=forestgreen,opacity=0.4] (1.75,-8.5) rectangle (3.25,-9);
        \draw [fill=forestgreen,opacity=0.8] (3.25,-8.5) rectangle (4.75,-9);
        \draw [fill=forestgreen,opacity=0.2] (4.75,-8.5) rectangle (6.25,-9);
        \draw [fill=forestgreen,opacity=0.1] (1.75,-9) rectangle (3.25,-9.5);
        \draw [fill=forestgreen,opacity=0.2] (3.25,-9) rectangle (4.75,-9.5);
        \draw [fill=forestgreen,opacity=0.3] (4.75,-9) rectangle (6.25,-9.5);

        \draw [fill=sepia,opacity=0.3] (10,-6) rectangle (14.5,-6.5);
        \draw [fill=orangered,opacity=0.3] (10,-6.5) rectangle (14.5,-7);
        \draw [gray] (11.5,-6) -- (11.5,-7);
        \draw [gray] (13,-6) -- (13,-7);
        \draw [fill=midnightblue,opacity=0.2] (10,-6) rectangle (14.5,-7);
        
        \draw [rounded corners=5pt,->] (9.5,-0.5) -- (12.25, -0.5) -- (12.25, -5.75);
        \draw [->] (6.25,-6.5) -- (9.4, -6.5);
        \node at (9.7,-6.5) {$\oplus$};
      \end{scope}

      \begin{scope}[shift={(-1.5,-19.5)}] 
        
        \draw [fill=sepia,opacity=0.1] (0,0) rectangle (4.5,0.5);
        \draw [fill=sepia,opacity=0.3] (4.5,0) rectangle (7.5,0.5);        
        \draw [fill=orangered,opacity=0.1] (0,0.5) rectangle (4.5,1);
        \draw [fill=orangered,opacity=0.3] (4.5,0.5) rectangle (7.5,1);

        \node at (0.75,-0.75) {\small \rotatebox{45}{[PAD]}}; 
        \node at (2.25,-0.75) {\small \rotatebox{45}{[PAD]}}; 
        \node at (3.75,-0.75) {\small \rotatebox{45}{[PAD]}}; 
        \node at (5.25,-0.75) {\small \rotatebox{45}{Match}}; 
        \node at (6.75,-0.75) {\small \rotatebox{45}{report}};

        \draw [gray] (0,0.5) -- (7.5,0.5);
        \draw [gray] (1.5,0) -- (1.5,1);
        \draw [gray] (3,0) -- (3,1);
        \draw [gray] (4.5,0) -- (4.5,1);
        \draw [gray] (6,0) -- (6,1);
        \node at (8.5,0.75) {$x'_i+p'_i$};
        \node at (8.5,0.25) {$t_D$};        
        \node at (-0.5,0.5) {$\mathbf{g}$};

        
        \node at (1,3) {Convolutions};

        \draw [fill=midnightblue,opacity=0.3] (3.75,1.5) -- (6.75, 1.5) -- (6.75, 3.5) -- (3.75,1.5);
        \draw [fill=midnightblue,opacity=0.2] (2.25,1.5) -- (5.25, 1.5) -- (5.25, 3.5) -- (2.25,1.5);
        \draw [fill=midnightblue,opacity=0.1] (0.75,1.5) -- (3.75, 1.5) -- (3.75, 3.5) -- (0.75,1.5);
        
        \draw [fill=white,opacity=1] (3,3.25) rectangle (4,3.75);
        \draw [fill=white,opacity=1] (4.5,3.25) rectangle (5.5,3.75);
        \draw [fill=white,opacity=1] (6,3.25) rectangle (7,3.75);
        
        \draw (3.5,3.25) -- (3.5,3.75);
        \draw (5,3.25) -- (5,3.75);
        \draw (6.5,3.25) -- (6.5,3.75);
        
        \node at (1,4.45) {GLU};

        \draw [->] (3.25,3.75) -- (3.25,4.25);
        \draw [gray] (3.25, 4.45) circle (0.2cm);
        \node at (3.25, 4.45) {$f$};
        \draw [->] (3.75,3.75) -- (3.75,5);
        \node at (3.75, 5.2) {$\otimes$};
        \draw [->] (3.25, 4.65) -- (3.25, 5.2) -- (3.5,5.2);
        \draw [->] (3.75,5.4) -- (3.75,6);

        \draw [->] (4.75,3.75) -- (4.75,4.25);
        \draw [gray] (4.75, 4.45) circle (0.2cm);
        \node at (4.75, 4.45) {$f$};
        \draw [->] (5.25,3.75) -- (5.25,5);
        \node at (5.25, 5.2) {$\otimes$};
        \draw [->] (4.75, 4.65) -- (4.75,5.2) -- (5,5.2);
        \draw [->] (5.25,5.4) -- (5.25,6);

        \draw [->] (6.25,3.75) -- (6.25,4.25);
        \draw [gray] (6.25, 4.45) circle (0.2cm);
        \node at (6.25, 4.45) {$f$};
        \draw [->] (6.75,3.75) -- (6.75,5);
        \node at (6.75, 5.2) {$\otimes$};
        \draw [->] (6.25, 4.65) -- (6.25,5.2) -- (6.5,5.2);
        \draw [->] (6.75,5.4) -- (6.75,6);

        \draw [fill=midnightblue,opacity=0.1] (3,6) rectangle (4.5,7);
        \draw [fill=midnightblue,opacity=0.2] (4.5,6) rectangle (6,7);
        \draw [fill=midnightblue,opacity=0.3] (6,6) rectangle (7.5,7);
        
        \node at (2.5,6.5) {$\mathbf{h^{\ell}}$};

        \draw [fill=forestgreen,opacity=0.3] (11.5,8) rectangle (13,9);
        \draw [fill=forestgreen,opacity=0.2] (13,8) rectangle (14.5,9);
        \draw [fill=forestgreen,opacity=0.3] (14.5,8) rectangle (16,9);     
        \draw [->] (12.25,8) -- (12.25,7);     
        \draw [->] (13.75,8) -- (13.75,7);     
        \draw [->] (15.25,8) -- (15.25,7); 
        \node at (11,8.5) {$\mathbf{c^{\ell}}$};

        \draw [fill=midnightblue!20!forestgreen,opacity=0.3] (11.5,6) rectangle (13,7);
        \draw [fill=midnightblue!60!forestgreen,opacity=0.2] (13,6) rectangle (14.5,7);
        \draw [fill=midnightblue!80!forestgreen,opacity=0.3] (14.5,6) rectangle (16,7);        
        \draw [->] (7.5,6.5) -- (10.9, 6.5);
        \node at (11.2,6.5) {$\oplus$}; 
        \node at (10.5,7) {$\mathbf{h^L}$};
        
        \draw [->] (12.25,6) -- (12.25,1); 
        \draw [fill=gray,opacity=0.3] (11.5,0) rectangle (13,1);
        \draw [->] (13.75,6) -- (13.75,1);     
        \draw [fill=gray,opacity=0.3] (13,0) rectangle (14.5,1);
        \draw [->] (15.25,6) -- (15.25,1);     
        \draw [fill=gray,opacity=0.3] (14.5,0) rectangle (16,1);
        
        \node at (12.25,-0.75) {\small \rotatebox{45}{Match}}; 
        \node at (13.75,-0.75) {\small \rotatebox{45}{report}}; 
        \node at (15.25,-0.75) {\small \rotatebox{45}{.}};

        \draw [rounded corners=5pt,->] (3.75,7) -- (3.75,7.5) -- (1.5, 7.5) -- (1.5, 11.25) -- (3.25, 11.25);
        \draw [rounded corners=5pt,->] (5.25,7) -- (5.25,8) -- (2, 8) -- (2, 10.75) -- (3.25, 10.75);
        \draw [rounded corners=5pt,->] (6.75,7) -- (6.75,8.5) -- (2.5, 8.5) -- (2.5, 10.25) -- (3.25, 10.25);
        \draw [gray] (12.25,11.25) circle (0.2cm);
        \node at (12.25,11.25) {\fontsize{3}{5}\selectfont $\sum$};
        \draw [->] (7.75, 11.25) -- (12.05,11.25);
        \draw [->] (12.25, 11.05) -- (12.25,9);

        \draw [gray] (13.75,10.75) circle (0.2cm);
        \node at (13.75,10.75) {\fontsize{3}{5}\selectfont $\sum$};
        \draw [->] (7.75, 10.75) -- (13.5,10.75);
        \draw [->] (13.75, 10.5) -- (13.75,9);
        
        \draw [gray] (15.25,10.25) circle (0.2cm);
        \node at (15.25,10.25) {\fontsize{3}{5}\selectfont $\sum$};
        \draw [->] (7.75, 10.25) -- (15.05,10.25);
        \draw [->] (15.25, 10.05) -- (15.25,9);
        
        \node at (15,11.75) {$\Big\Downarrow$};
      \end{scope}
    \end{tikzpicture}
  }
  \caption{Topic-conditioned convolutional model for extreme
    summarization. The input document ``England won.'' is encoded
    (top) using word ($x_i$) and position ($p_i$) embeddings, and word
    ($t'_i$) and document ($t_D$) topic vectors. At each time step of
    the decoding, a context representation (bottom left) is learned
    using word ($x'_i$) and position ($p'_i$) embeddings and the
    document ($t_D$) topic vector. Attention (center) is the dot
    product between decoder context representations and encoder
    representations. The conditional inputs~$c^l$ computed by the
    attention are added to the decoder states~$h^l$ which then predict
    the target words (bottom right). GLU stands for Gated Linear
    Units. Both encoder and decoder pad input and output sequences
    respectively to match the receptive width of the convolutional
    filters. Here, the decoder generates summary ``Match
    report.''.}\label{fig:architecture}
\end{figure}
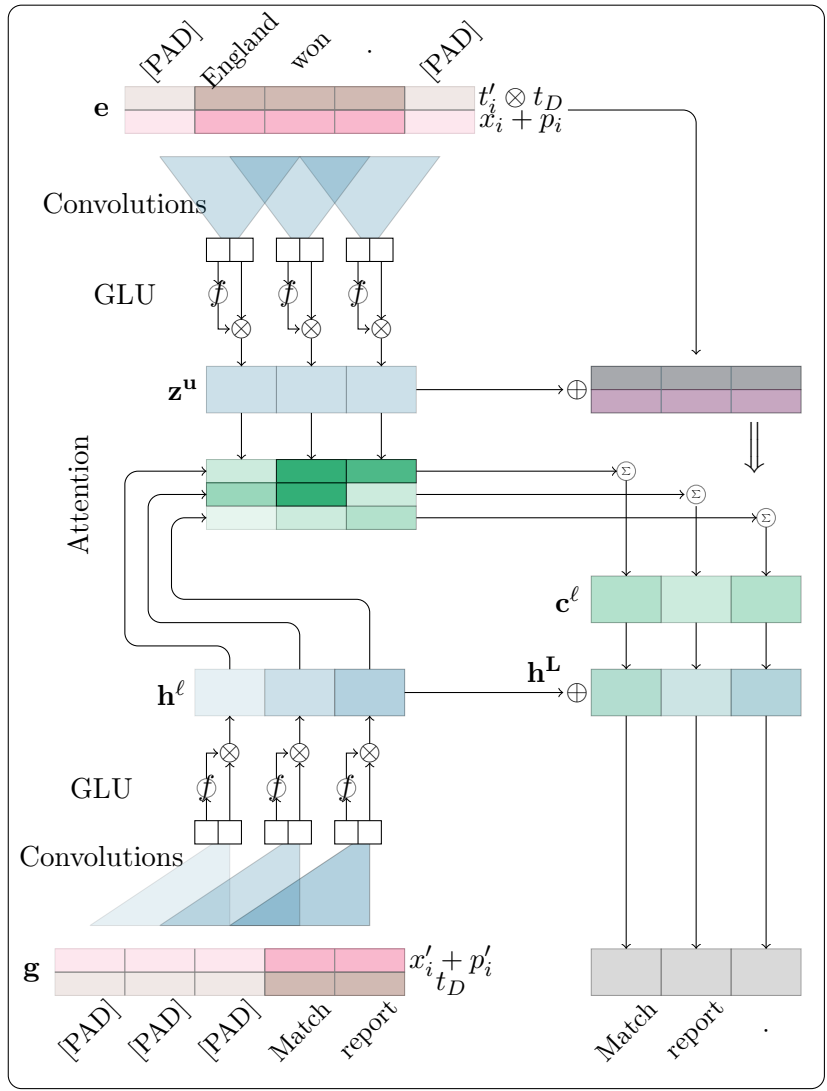

\subsection{Model Overview}

At the core of our model is a simple convolutional block structure
that computes intermediate states based on a fixed number of input
elements. Our convolutional encoder (shown at the top of Figure
\ref{fig:architecture}) applies this unit across the document. We
repeat these operations in a stacked fashion to get a multi-layer
hierarchical representation over the input document where words at
closer distances interact at lower layers while distant words interact
at higher layers. The interaction between words through hierarchical
layers effectively captures long-range dependencies.

Analogously, our convolutional decoder (shown at the bottom of
Figure~\ref{fig:architecture}) uses the multi-layer convolutional
structure to build a hierarchical representation over what has been
predicted so far. Each layer on the decoder side determines useful
source context by attending to the encoder representation before it
passes its output to the next layer. This way the model remembers
which words it previously attended to and applies multi-hop attention
(shown in the middle of Figure~\ref{fig:architecture}) per time
step. The output of the top layer is passed to a softmax classifier to
predict a distribution over the target vocabulary.

Our model assumes access to word and document topic
distributions. These can be obtained by any topic model, however we
use Latent Dirichlet Allocation (LDA; \citealt{Blei:2003:LDA}) in our
experiments; we pass the distributions obtained from LDA directly to
the network as additional input. This allows us to take advantage of
topic modeling without interfering with the computational advantages
of the convolutional architecture. 


\subsection{Topic Sensitive Embeddings}

Let~$D$ denote a document consisting of a sequence of words $(w_1,
\ldots, w_m)$; we embed~$D$ into a distributional space $\mathbf{x} =
(x_1, \ldots, x_m)$ where $x_i \in \mathbb{R}^f$ is a column in
embedding matrix $M \in \mathbb{R}^{V\times f}$ (where $V$ is the
vocabulary size). We also embed the absolute word positions in the
document \mbox{$\mathbf{p} = (p_1, \ldots, p_m)$} where $p_i \in
\mathbb{R}^f$ is a column in position matrix $P \in
\mathbb{R}^{N\times f}$, and $N$ is the maximum number of positions;
$p_i$ is the position embedding of word $w_i$ at position~$i$ in the
input sequence. Position
embeddings 
have proved useful for convolutional sequence modeling
\cite{convseq2seq}, because, in contrast to RNNs, they do not observe
the temporal positions of words \cite{D16-1248}. Let $t_D \in
\mathbb{R}^{f'}$ be the topic distribution of document~$D$ and
$\mathbf{t'} = (t'_1, \ldots, t'_m)$ the topic distributions of words
in the document (where $t'_i \in \mathbb{R}^{f'}$). During encoding,
we represent document~$D$ via $\mathbf{e} = (e_1, \ldots, e_m)$,
where~$e_i$ is:
\begin{align}
e_i = [(x_i+p_i);(t'_i\otimes t_D)] \in \mathbb{R}^{f+f'}, \label{eq:encoder}
\end{align}
\noindent and $\otimes$ denotes point-wise multiplication. The topic
distribution $t'_i$ of word $w_i$ essentially captures how topical the
word is in itself (local context), whereas the topic
distribution~$t_D$ represents the overall theme of the document
(global context). The encoder essentially enriches the context of the
word with its topical relevance to the document.

For every output prediction, the decoder estimates representation
$\mathbf{g} = (g_1, \ldots, g_n)$ for previously predicted words
$(w'_1, \ldots, w'_n)$ where $g_i$ is:
\begin{align}
g_i = [(x'_i+p'_i);t_D] \in \mathbb{R}^{f+f'}, \label{eq:decoder}
\end{align}
\noindent $x'_i$ and $p'_i$ are word and position embeddings of
previously predicted word~$w'_i$, and $t_D$ is the topic distribution
of the input document. Note that the decoder does not use the topic
distribution of~$w'_i$ as computing it on the fly would be
expensive. However, every word prediction is conditioned on the topic
of the document, enforcing the summary to have the same theme as the
document.


\subsection{Multi-layer Convolutional Structure}

Each convolution block, parametrized by $W \in \mathbb{R}^{2d\times
  kd}$ and $b_w \in \mathbb{R}^{2d}$, takes as input $X \in
\mathbb{R}^{k\times d}$ which is the concatenation of $k$ adjacent
elements embedded in a~$d$ dimensional space, applies one dimensional
convolution and returns an output element $Y \in \mathbb{R}^{2d}$. We
apply Gated Linear Units (GLU, $v:\mathbb{R}^{2d} \rightarrow
\mathbb{R}^{d}$, \citeR{pmlr-v70-dauphin17a}) on the output of
convolution~$Y$. Subsequent layers operate over the $k$ output
elements of the previous layer and are connected through residual
connections \cite{He2016DeepRL} to allow for deeper hierarchical
representation. We denote the output of the $\ell$th layer as
$\mathbf{h^{\ell}} = (h^{\ell}_1, \ldots, h^{\ell}_n)$ for the decoder
network, and $\mathbf{z^{\ell}} = (z^{\ell}_1, \ldots, z^{\ell}_m)$
for the encoder network.

\subsection{Multi-hop Attention}

Our encoder and decoder are tied via a multi-hop attention
mechanism. For each decoder layer $\ell$, we compute the attention
$a^{\ell}_{ij}$ of state $i$ and source element $j$ as:

\begin{align}
a^{\ell}_{ij} = \frac{\mbox{exp}(d^{\ell}_i \cdot z^u_j)}{\sum^m_{t=1} \mbox{exp}(d^{\ell}_i \cdot z^u_t)}, \label{eq:attention}
\end{align}

\noindent where $d^{\ell}_i = W^{\ell}_dh^{\ell}_i+b^{\ell}_i+g_i$ is
the decoder state summary combining the current decoder state
$h^{\ell}_i$ and the previous output element embedding
$g_i$. Vector~$\mathbf{z^u}$ is the output from the last encoder layer
$u$. The conditional input $c^{\ell}_i$ to the current decoder layer
is a weighted sum of the encoder outputs as well as the input element
embeddings $e_j$:

\begin{align}
c^{\ell}_i = \sum^m_{j=1} a^{\ell}_{ij}(z^u_j+e_j). \label{eq:decinput}
\end{align}

\noindent The attention mechanism described here performs multiple
attention ``hops'' per time step and considers which words have been
previously attended to. It is therefore different from single-step
attention in recurrent neural networks \cite{bahdanau-arxiv14}, where
the attention and weighted sum are computed over $\mathbf{z^u}$ only.

Our network uses multiple linear layers to project between the
embedding size~$(f+f')$ and the convolution output size~$2d$. These
are applied to~$\mathbf{e}$ (before feeding it to the encoder), to the
final encoder output $\mathbf{z^u}$, to all decoder
layers~$\mathbf{h^{\ell}}$ (for the attention score computation), and
to the final decoder output~$\mathbf{h^L}$ (before the softmax).  We
pad the input with~$k-1$ zero vectors on both left and right sides to
ensure that the output of the convolutional layers matches the input
length. During decoding, we ensure that the decoder does not have
access to future information; we start with $k$~zero vectors and shift
the convolutional block to the right after every prediction. The final
decoder output $\mathbf{h^L}$ is used to compute the distribution over
the target vocabulary as:

\begin{align}
  p(y_{i+1}|y_1,\ldots,y_i,D,t_D,\mathbf{t'}) = \mbox{softmax}(W_oh^L_i+b_o) \in \mathbb{R}^T \label{eq:prediction}
\end{align}

\noindent where, $W_o$ and $b_o$ are the parameters of the softmax
layer and $T$ is the size of the target vocabulary. We also use layer
normalization and weight initialization to stabilize learning. We use
cross-entropy loss to maximize the likelihood of the ground-truth
sequence $(y^{\ast}_1, \ldots, y^{\ast}_n)$:

\begin{align}
  L(\mathbf{\theta}) 
  = \sum_{i=0}^{n-1} \mbox{log } p(y^{\ast}_{i+1}|y^{\ast}_1,\ldots,y^{\ast}_i,D,t_D,\mathbf{t'},\mathbf{\theta}) 
 \label{eq:loss}
\end{align}

\noindent Our topic-enhanced model calibrates long-range dependencies
with globally salient content. As a result, it provides a better
alternative to vanilla convolutional sequence models
\cite{convseq2seq} and RNN-based summarization models \cite{see-acl17}
for capturing cross-document inferences and paraphrasing.  At the same
time it retains the computational advantages of convolutional
models. Each convolution block operates over a fixed-size window of
the input sequence, allowing for simultaneous encoding of the input
and ease in learning due to the fixed number of non-linearities and
transformations for words in the input sequence.


\section{Experimental Setup}
\label{sec:setup}

In this  section we present  our experimental setup for  assessing the
performance   of   our   \textbf{T}opic-aware   \textbf{Conv}olutional
\textbf{S}equence   \textbf{to}  \textbf{S}equence   model  which   we
abbreviate to \textsc{T-ConvS2S}.  We evaluate  our model on our newly
collected  XSum dataset  and  show  that it  is  suitable for  extreme
summarization. We also report experiments on the abstractive subset of
the Newsroom dataset  (Newsroom-Abs; \citeR{newsroom-naacl18}). In the
following, we discuss implementation details, present the systems used
for comparison  with our approach,  and explain how system  output was
evaluated.

\subsection{Comparison Systems} 
\label{subsec:comparisons}


We report results with various systems which were all trained on the
XSum dataset to generate a one-line summary given an input news
article.  We compared \mbox{\textsc{T-ConvS2S}} against three
extractive systems: a baseline which randomly selects a sentence from
the input document (\textsc{random}), a baseline which simply selects
the leading sentence from the document (\textsc{lead}), and an oracle
which selects a single-best sentence in each document
(\textsc{ext-oracle}). The latter is often used as an upper bound for
extractive methods. We also compared our model against the RNN-based
abstractive systems introduced in \newcite{see-acl17}.\footnote{
  State-of-the-art abstractive systems on the CNN/Daily mail and New
  York Times datasets
  \cite{asli-multiagent18,Pasunuru-multireward18,kryciski-EtAl:2018:EMNLP}
  use reinforcement learning to directly optimize the evaluation
  metric relevant for the summarization task. Although our model could
  be optimized with reinforcement learning objectives, we leave this
  to future work and present comparisons with related models which are
  all trained with the maximum-likelihood objective.} In particular,
we experimented with an attention-based sequence-to-sequence model
(\textsc{Seq2Seq}), a pointer-generator model which allows us to copy
words from the source text (\textsc{PtGen}), and a pointer-generator
model with a coverage mechanism to keep track of words that have been
summarized (\textsc{PtGen+Covg}). Finally, we compared our model
against two convolutional abstractive systems: the vanilla convolution
sequence-to-sequence model of \citeA{convseq2seq} (\textsc{ConvS2S})
and a variant thereof augmented a copy mechanism
(\textsc{ConvS2S+Copy}) which copies words from the input document via
pointing \cite{vinyals-nips15,see-acl17} while retaining the ability
to produce novel words from a fixed
vocabulary.\footnote{\textsc{ConvS2S+Copy} estimates the generation
  probability $p_{gen} \in [0,1]$ at each decoding step~$i$ as:
  $$p_{gen} = \sigma(w_h h_i^L + w_c c_i^L + w_g g_i +
  b_{gen})$$
  \noindent using the final decoder state $h_i^L$, the final context
  vector $c_i^L$ and the decoder input $g_i$. $L$ is the final layer
  of the decoder. $w_h$, $w_c$, $w_g$ and $b_{gen}$ are model
  parameters and $\sigma$ is the non-linear sigmoid function. We
  estimate the final probability distribution $p'(w) \in R^{T'}$ over
  the extended vocabulary~$T'$ denoting the union of the target
  vocabulary~$T$ and the words of the input document as
  $$p'(w) = p_{gen} {p(w)}^{\alpha} + (1-p_{gen}) \sum_{j:w_j=w}
  (a_{ij}^L)^{\beta},$$
  \noindent where~$p(w)$ is the target vocabulary distribution
  (estimated using Equation~\eqref{eq:prediction} and $a_{ij}^L$ is
  the attention for the final decoder layer $L$ (estimated using
  Equation~\eqref{eq:attention}). $\alpha$ and $\beta$ are scaling
  parameters (used to stabilize convolutional learning) estimated as
  $\frac{\log(|T'|)}{\log(|T|)}$ and
  $\frac{\log(|T'|)}{\log(\mathrm{EncLen})}$, respectively (where
  $\mathrm{EncLen}$ is the encoder length). \textsc{ConvS2S+Copy} uses
  $p_{gen}$ to switch between generating a novel word from a fixed
  vocabulary by sampling from $p(w)$ or copying a word from the source
  text by sampling from the attention distribution $a_{ij}^L$.}

For our experiments on Newsroom-Abs, we again compared
\mbox{\textsc{T-ConvS2S}} against the extractive systems
\textsc{random}, \textsc{lead} and \textsc{ext-oracle}, the recurrent
abstractive systems \textsc{Seq2Seq}, \textsc{PtGen}, and
\textsc{PtGen+Covg}, and the convolutional systems \textsc{ConvS2S}
and \textsc{ConvS2S+Copy}.
All systems were retrained on the Newsroom-Abs training set.
\textsc{lead} selects the first 2 sentences to form the summary while
\textsc{random} selects random 2 sentences from the input document to
form the summary. \textsc{ext-oracle} creates an oracle summary by
selecting the best possible set of sentences in the document that
gives the highest ROUGE \cite{rouge} with respect to the gold summary.

\begin{table}[t!]
  \center{\footnotesize 
    \begin{tabular}{@{}l@{\hspace{1.9ex}}p{13.5cm} }
      \thickhline 
      \multicolumn{2}{c}{XSum documents}\\
      \hline 
      {T1:}& murder, charge, court, police, arrest, guilty, sentence, boy, bail, space, crown, trial \\
      
      {T2:}& abuse, church, bishop, child, catholic, gay, pope, school, christian, priest, cardinal \\
      
      {T3:}& council, people, government, local, housing, home, house, property, city, plan, authority \\
      
      {T4:}& party, clinton, trump, climate, poll, vote, plaid, election, debate, change, candidate, campaign\\
      
      {T5:}& country, growth, report, business, export, fall, bank, security, economy, rise, global, inflation\\
      
      {T6:}& hospital, patient, trust, nhs, people, care, health, service, staff, report, review, system, child\\
      \thickhline
      \multicolumn{2}{c}{Newsroom Abstractive documents}\\
      \hline 
      {T1:}& fund, investment, firm, asset, capital, financial, corporate, management, return, profit, equity \\
      
      {T2:}& building, design, build, square, space, office, architect, center, architecture, interior, project \\    
  
      {T3:}& award, parade, beverly, actress, annual, hills, star, red, hollywood, carpet, premiere, pose \\
      
      {T4:}& company, business, customer, industry, consumer, service, product, revenue, fortune, startup \\
      
      {T5:}& military, force, afghanistan, government, security, troops, war, country, taliban, attack, army \\
      
      {T6:}& party, government, minister, leader, political, prime, election, vote, power, country, parliament \\ 
      \thickhline
    \end{tabular}     
  }
  \caption{Example topics learned by an LDA model on  XSum and Newsroom documents (training portion).}\label{fig:lda-topics}
\end{table}

\subsection{Model Parameters and Optimization} 

We did not anonymize entities but worked with a lowercased version of
the XSum and Newsroom-Abs datasets. During training and at test time
input documents were truncated to 400~tokens and the length of the
summary was limited to 90~tokens.

We trained two separate LDA models \cite{Blei:2003:LDA} on XSum and
Newsroom documents (training portion). We therefore obtained for each
word a probability distribution over topics which we used to estimate
$\mathbf{t'}$; the topic distribution $t_D$ can be inferred for any
new document, at training and test time. We explored several LDA
configurations on held-out data, and obtained best results with~512
topics for XSum and~256 topics for Newsroom.  LDA models were trained
with $\alpha$\footnote{$\alpha$ controls the prior distribution over
  topics for individual documents.}  set to a fixed normalized
asymmetric prior of~$1/\mbox{number of topics}$; we let
the model learn an asymmetric prior~$\eta$\footnote{$\eta$ controls
  the prior distribution over words for individual topics.}  from the
data. Table~\ref{fig:lda-topics} shows some of the topics learned by
the LDA models.\footnote{We used a multi-core implementation of LDA made
  available by \texttt{gensim} at
  \url{https://radimrehurek.com/gensim/models/ldamulticore.html}.}


For all RNN-based models\footnote{We used the code available at
  \url{https://github.com/abisee/pointer-generator}.}
(\textsc{Seq2Seq}, \textsc{PtGen}, and \textsc{PtGen+Covg}) we used
the best settings reported on the CNN and DailyMail data
\cite{see-acl17} All models had 256~dimensional hidden states and
128~dimensional word embeddings. They were trained using Adagrad
\cite{Duchi:2011} with learning rate set to~0.15 and an initial
accumulator value of~0.1. We used gradient clipping with a maximum
gradient norm of~2, without any regularization and the loss on the
validation set to implement early stopping. All models trained on the
XSum and Newsroom datasets have the same settings.


For all convolutional models\footnote{We used the code available at
  \url{https://github.com/facebookresearch/fairseq-py}.}
(\textsc{ConvS2S}, \textsc{ConvS2S+Copy}, and \textsc{T-ConvS2S}) we
used 512 dimensional hidden states, word embeddings and position
embeddings for XSum and 256 dimensional hidden states, word embeddings
and position embeddings for Newsroom. All models were trained with
Nesterov's accelerated gradient method \cite{Sutskever:2013} using a
momentum value of~0.99 and renormalized gradients if their norm
exceeded~0.1 \cite{Pascanu:2013}. We used a learning rate of~0.10 for
\textsc{ConvS2S} and \textsc{T-ConvS2S}, and~0.02 for
\textsc{ConvS2S+Copy}.\footnote{\textsc{ConvS2S+Copy} failed to
  converge with learning rate greater than~0.02.} Once the
validation perplexity stopped improving, we reduced the learning rate
by an order of magnitude after each epoch until it fell
below~$10^{-4}$. We also applied a dropout of~0.2 to the embeddings,
the decoder outputs and the input of the convolutional
blocks. Gradients were normalized by the number of non-padding tokens
per mini-batch. We also used layer normalization and weight
normalization for all layers except for lookup tables to stabilize
learning.


All neural models, including ours and those based on RNNs
\cite{see-acl17}, had a vocabulary of 50,000 words and were trained on
a single Nvidia M40 GPU with a batch size of~32 sentences.  Summaries
at test time were obtained using beam search (with beam size~10) in
all cases.

\subsection{Evaluation}

We evaluated summarization quality automatically using F$_1$
$\mbox{ROUGE}$ \cite{rouge}. Unigram and bigram overlap
($\mbox{ROUGE-1}$ and $\mbox{ROUGE-2}$) are a proxy for assessing
informativeness and the longest common subsequence ($\mbox{ROUGE-L}$)
represents fluency.\footnote{We used \texttt{pyrouge} to compute all
  ROUGE scores, with parameters ``-a -c 95 -m -n 4 -w 1.2.''}  In
addition to ROUGE which can be misleading when used as the only means
to assess the informativeness of summaries
\cite{schluter:2017:EACLshort,hardy-etal-acl19}, we also evaluated system output by
eliciting human judgments in two ways. 


In our first experiment, participants were asked to compare summaries
produced by different systems. The study was conducted on the Amazon
Mechanical Turk platform using \textit{Best-Worst Scaling} (BWS;
\citeR{louviere1991best}; \citeR{louviere2015best}), a less
labor-intensive alternative to paired comparisons that has been shown
to produce more reliable results than rating scales
\cite{bestworstscaling}. Participants were presented with a document
and summaries generated from three systems and were asked to decide
which summary was the \textit{best} and which one was the
\textit{worst} in order of informativeness (does the summary capture
important information in the document?) and fluency (is the summary
written in well-formed English?). In two separate studies, we randomly
selected 50~documents from the XSum and Newsroom-Abs test set.  We
compared all possible system pairs for each document and collected
judgments from three different participants for each comparison. The
order of summaries was randomized per document and the order of
documents per participant. The score of a system was computed as the
percentage of times it was chosen as best minus the percentage of
times it was selected as worst. The scores range from -1 (worst) to 1
(best). Figures~\ref{fig:erroranalysis-xsum}
and~\ref{fig:erroranalysis-newsroom} show example summaries from the
XSum and Newsroom datasets used for this study.


For our second experiment we used a question-answering (QA) paradigm
\cite{Clarke:Lapata:2010,narayan-rank18} to assess the degree to which
the models retain key information from the document. We wrote
fact-based questions for each document, just by reading the reference
summary, under the assumption that it highlights the most important
content of the news article. Questions were formulated so as not to
reveal answers to subsequent questions. Participants read the output
summaries and answered the questions as best they could without access
to the document or the gold summary. The more questions can be
answered, the better the corresponding system is at summarizing the
document as a whole.  Five participants answered questions for each
summary. Answers again were elicited using Amazon's Mechanical Turk
crowdsourcing platform. We uploaded the data in batches (one system at
a time) to ensure that the same participant does not evaluate
summaries from different systems on the same set of questions.  We
followed the scoring mechanism introduced in
\newcite{Clarke:Lapata:2010}. A correct answer was marked with a score
of one, partially correct answers with a score of~0.5, and zero
otherwise. The final score for a system is the average of all its
question scores.

We used the same 100 documents (50 documents for XSum and 50 documents
for Newsroom) as in our first elicitation study. For XSum, we created
100~questions in total; we wrote two fact-based questions per
document.  For Newsroom
summaries, we were often not able to write more than one fact-based
questions per document. Consequently, we only have 61~questions in
total. Figures~\ref{fig:erroranalysis-xsum}
and~\ref{fig:erroranalysis-newsroom} show example summaries and their
corresponding questions for XSum and Newsroom, respectively.


\section{Results}
\label{sec:results}

In this section we present results for our model and comparison
systems on the XSum dataset; we also discuss experiments on Newsroom
\cite{newsroom-naacl18} and analyze quantitative and qualitative
differences between the two datasets.

\subsection{Results on the XSum Dataset}
\label{sec:results-xsum}

\begin{table}[t]
  \begin{center}{ 
  \begin{tabular}{ l | c c c }
    \thickhline
    Models & R1 & R2 & RL \\ \thickhline
    Random & 15.16 & 1.78 & 11.27 \\ 
    \textsc{lead} & 16.30 & 1.60 & 11.95 \\ 
    \textsc{ext-oracle} & 29.79 & 8.81 & 22.66 \\ \hline\hline
    \textsc{Seq2Seq} &  28.42 & 8.77 &  22.48 \\
    \textsc{PtGen} & 29.70 & 9.21 & 23.24 \\ 
    \textsc{PtGen+Covg} & 28.10 & 8.02 & 21.72 \\ 
    \hline\hline
    \textsc{ConvS2S} & 31.27 & 11.07 & 25.23 \\  
    \textsc{ConvS2S+Copy} & 29.80 & 10.10 & 24.10 \\ 
    \textsc{T-ConvS2S} (enc$_{t'}$) & 31.71 & 11.38 & 25.56 \\
    \textsc{T-ConvS2S} (enc$_{(t',t_D)}$) & 31.61 & 11.30 & 25.51 \\
    \textsc{T-ConvS2S} (enc$_{t'}$, dec$_{t_D}$) & 31.71 & 11.34 & 25.61 \\
    \textsc{T-ConvS2S} (enc$_{(t',t_D)}$, dec$_{t_D}$)  & {31.89} & {11.54} & {25.75} \\ 

    \thickhline  

  \end{tabular}
  }\end{center}
\caption{ROUGE results on XSum test set. We report \mbox{ROUGE-1}
  (R1), ROUGE-2 (R2), and ROUGE-L (RL) F$_1$ scores. Extractive
  systems are in the upper block, RNN-based
  abstractive  models are in
  the middle block, and convolutional systems are in the
  bottom block. \label{tab:rouge-xsum}}
\end{table}

\paragraph{Automatic Evaluation} 

Table~\ref{tab:rouge-xsum} summarizes our ROUGE-based results.  As can
be seen, \textsc{Seq2Seq} outperforms the \textsc{lead} and
\textsc{random} baselines by a large margin. \textsc{PtGen}, a
\textsc{Seq2Seq} model with a ``copying'' mechanism outperforms
\textsc{ext-oracle}, a ``perfect'' extractive system on ROUGE-2 and
ROUGE-L. This is in sharp contrast to the performance of these models
on the CNN/DailyMail \cite{see-acl17} and Newsroom datasets
\cite{newsroom-naacl18}, where they fail to outperform the
\textsc{lead}. The result provides further evidence that XSum is a
good testbed for abstractive summarization. \textsc{PtGen+Covg}, the
best performing abstractive system on the CNN/DailyMail datasets, does
not do well.  We believe that the coverage mechanism is more useful
when generating multi-line summaries and is basically redundant for
extreme summarization.

\textsc{ConvS2S}, the convolutional variant of \textsc{Seq2Seq},
significantly outperforms all \mbox{RNN-based}
abstractive systems.\footnote{Statistical
  significance at the~95\% confidence level is estimated using
  bootstrap resampling \cite{Davison:Hinkley:1997} with the official
  ROUGE script.}  We hypothesize that its superior performance stems
from the ability to better represent document content (i.e.,~by
capturing long-range dependencies). Surprisingly,
\textsc{ConvS2S+Copy}, a \textsc{ConvS2S} enhanced with a ``copying''
mechanism obtains performance inferior to \textsc{ConvS2S}.  Our
analysis revealed that the multi-hop attention mechanism of
\textsc{ConvS2S} is very effective in resolving the unknown (UNK)
words, by simply copying the most attended word $w_j$ (estimated using
the average attention scores as $\mathrm{argmax}_{w_j} \sum_{l=1}^L
a_{ij}^L/L)$) from the source text to replace an UNK word. The copy
mechanism unnecessarily over-parametrizes the \textsc{ConvS2S} model
leading to a drop in performance. For example, \textsc{ConvS2S}
correctly resolves two subsequent UNK words to ``Dick Advocaat''
whereas \textsc{ConvS2S+Copy} incorrectly resolves them to ``Dick
Dick'' as shown at the bottom of
Figure~\ref{fig:erroranalysis-xsum}. The \textsc{Seq2Seq} model
without the copy mechanism is prone to generating random rare words
(e.g., ``Andre Mccormack'' for the same example in
Figure~\ref{fig:erroranalysis-xsum}) or unresolved UNK words (see the
top example in Figure~\ref{fig:erroranalysis-newsroom}).\footnote{None
  of the models of \newcite{see-acl17} resolves UNK by simply copying
  the most attended word from the source text. \textsc{PtGen} and
  \textsc{PtGen+Covg} rely on the copy mechanism to sample words from
  the extended target vocabulary, including source
  words.\label{footnote:see-copy}}, while \textsc{PtGen} guides the
model towards sampling words from the source text.

Table~\ref{tab:rouge-xsum} also shows several variants of
\textsc{T-ConvS2S} including an encoder network enriched with
information about how topical a word is on its own (enc$_{t'}$) or in
the document (enc$_{(t',t_D)}$).  We also experimented with various
decoders by conditioning every prediction on the topic of the
document, basically encouraging the summary to be in the same theme as
the document (dec$_{t_D}$) or letting the decoder decide the theme of
the summary. Interestingly, all four \textsc{T-ConvS2S} variants
outperform \textsc{ConvS2S} and
\textsc{ConvS2S+Copy}. \textsc{T-ConvS2S} performs best when both
encoder and decoder are constrained by the document topic
(enc$_{(t',t_D)}$,dec$_{t_D}$). In the remainder of the paper, we
refer to this variant as \textsc{T-ConvS2S}.

\begin{figure*}[t!]
  \center{\footnotesize 
    \begin{tabular}{l p{9.5cm} l}
      \thickhline 
      

      \textsc{lead} & The Richmond Park and North Kingston MP said he was ``honoured'' after winning 70\% of the 9,227 votes cast using an online primary system. & [9.8, 0.0, 9.8] \\
      \textsc{ext-oracle} & Caroline Pidgeon is the Lib Dem candidate, Sian Berry will contest the election for the Greens and UKIP has chosen its culture spokesman Peter Whittle. &[34.1, 20.5, 34.1]\\
      {\textsc{Seq2Seq}} & \textcolor{midnightblue}{Zac Goldsmith} has been \textcolor{midnightblue}{re-elected as the Conservative candidate} for \textcolor{red}{the general election in London}, the party has announced. & [61.1, 23.5, 38.9] \\
      {\textsc{PtGen}} & UKIP leader \textcolor{red}{Nigel Goldsmith} has been \textcolor{red}{elected} as the \textcolor{red}{new mayor of London} to elect a new Conservative MP. &[45.7, 6.1, 28.6]  \\
      {\textsc{PtGen+Covg}} & \textcolor{midnightblue}{Zac Goldsmith} has been \textcolor{red}{re-elected as the Liberal Democrat candidate} for the \textcolor{red}{Labour party conference} in Richmond, North Wales. & [44.4, 17.7, 27.8] \\
      {\textsc{ConvS2S}}&  London mayoral candidate \textcolor{midnightblue}{Zac Goldsmith} has been \textcolor{red}{elected} as the \textcolor{red}{new mayor of London}.& [53.3, 21.4, 26.7]  \\
      {\textsc{ConvS2S+Copy}} & \textcolor{midnightblue}{Zac Goldsmith} has been \textcolor{red}{elected} mayor of \textcolor{midnightblue}{London's mayoral election}. & [51.9, 24.0, 37.0] \\
      {\textsc{T-ConvS2S}}& Former London mayoral candidate \textcolor{midnightblue}{Zac Goldsmith} has been \textcolor{midnightblue}{chosen to stand} in \textcolor{midnightblue}{the London mayoral election}. &[50.0, 26.7, 37.5] \\
      {\textsc{gold}}& \multicolumn{2}{p{12cm}}{\textcolor{midnightblue}{Zac Goldsmith} will \textcolor{midnightblue}{contest} the 2016 \textcolor{midnightblue}{London mayoral election} for the Conservatives, it has been announced.} \\
      {Questions}& \multicolumn{2}{l}{(1) {Who will \textcolor{midnightblue}{contest} for the Conservatives?} (\textcolor{midnightblue}{Zac Goldsmith})}\\
      &  \multicolumn{2}{l}{(2) {For what election will he/she contest?} (\textcolor{midnightblue}{The London mayoral election})} \\
            
      \hline

      \textsc{lead} & The 68-year-old Dutchman was appointed in March, when the black cats were one point above the relegation zone. & [15.4, 0.0, 15.4] \\
      {\textsc{ext-oracle}} & North-east rivals Newcastle are the only team below them in the Premier League table. &[35.3, 18.8, 35.3] \\
      {\textsc{Seq2Seq}} & Sunderland ladies have confirmed that former Sunderland striker \textcolor{red}{Andre Mccormack} has been \textcolor{red}{sacked} as the new \textcolor{midnightblue}{manager of} Championship Club \textcolor{midnightblue}{Sunderland}. & [20.0, 0.0, 15.0] \\
      {\textsc{PtGen}} &Sunderland have \textcolor{red}{appointed} former \textcolor{midnightblue}{Sunderland boss} \textcolor{midnightblue}{Dick Advocaat} as manager at the end of the season to sign a new deal. &[45.0, 10.5, 30.0] \\
      {\textsc{PtGen+Covg}} & Sunderland have \textcolor{red}{sacked} \textcolor{midnightblue}{manager} \textcolor{midnightblue}{Dick Advocaat} after eight games in charge of the Club's Premier League rivals Sunderland. & [36.8, 11.1, 31.6] \\
      {\textsc{ConvS2S}} &Sunderland have \textcolor{red}{sacked} \textcolor{midnightblue}{manager} \textcolor{midnightblue}{Dick Advocaat} after less than three months in charge. &[25.0, 6.7, 18.8] \\
      {\textsc{ConvS2S+Copy}} & Sunderland have \textcolor{red}{appointed Dick Dick} as their new \textcolor{midnightblue}{manager} on a three-year deal. & [18.2, 0.00, 12.1] \\
      {\textsc{T-ConvS2S}}& \textcolor{midnightblue}{Dick Advocaat} has \textcolor{midnightblue}{resigned} as \textcolor{midnightblue}{Sunderland manager} until the end of the season.& [56.3, 33.3, 56.3] \\
      {\textsc{gold}} &\multicolumn{2}{p{12cm}}{\textcolor{midnightblue}{Dick Advocaat} has \textcolor{midnightblue}{resigned} as \textcolor{midnightblue}{Sunderland boss}, with the team yet to win in the Premier League this season.} \\
      Questions& (1) {Who has \textcolor{midnightblue}{resigned}?}
      (\textcolor{midnightblue}{Dick Advocaat}) \\& (2) {From what post has he/she resigned?} (\textcolor{midnightblue}{Sunderland boss}) \\

      \thickhline

    \end{tabular}     
  }
  \caption{Example output summaries on the XSum test set with
    [ROUGE-1, ROUGE-2 and ROUGE-L] scores, goldstandard reference, and
    corresponding questions. Words highlighted in blue are either the
    right answer or constitute appropriate context for inferring it;
    words in red lead to the wrong answer.
  }\label{fig:erroranalysis-xsum}
\end{figure*}

\begin{table}[t]
  \begin{center}{ 
  \begin{tabular}{ l | c c c c } 
    \thickhline
    \multirow{2}{*}{Models} & \multicolumn{4}{c}{\% of novel n-grams in generated summaries}  \\
    & unigrams & bigrams & trigrams & 4-grams  \\ \thickhline 
    \textsc{Seq2Seq} & {36.66} & {82.17} & {95.58} & {98.63}  \\
    \textsc{PtGen} & 27.40 & 73.33 & 90.43 & 96.04 \\ 
    \textsc{PtGen+Covg} & 25.71 & 70.76 & 88.87 & 95.24 \\ \hline\hline
    \textsc{ConvS2S} & 31.28 & 79.50 & 94.28 & 98.10 \\
    \textsc{ConvS2S+Copy} & 32.30 & 79.56 & 94.46 & 98.21\\
    \textsc{T-ConvS2S}  & 30.73 & 79.18 & 94.10 & 98.03 \\ \hline\hline
    \textsc{gold} & 35.76 & 83.45  & 95.50  & 98.49 \\ \thickhline
  \end{tabular}}
  \end{center}
  \caption{Proportion of novel $n$-grams in summaries generated by various models on the XSum test set.
    \label{tab:novelngram-xsum-models}}
\end{table}

\paragraph{How Abstractive are the Generated Summaries?}

We further assessed the extent to which various models are able to
perform rewriting by generating genuinely abstractive
summaries. Table~\ref{tab:novelngram-xsum-models} shows the proportion
of novel $n$-grams for abstractive systems based on RNNs
(\textsc{Seq2Seq}, \textsc{PtGen} and \textsc{PtGen+Covg}) and our
convolutional models (\textsc{ConvS2S}, \textsc{ConvS2S+Copy} and
\textsc{T-ConvS2S}). We omit extractive systems (\textsc{lead} and
\textsc{ext-oracle}) as they are not capable of generating summaries
from scratch with novel $n$-grams.

Overall, we observe that all abstractive models generate a fair amount
of novel constructions that go beyond what is said in the source
document. This result further supports our claim that XSum is an
appropriate testbed for abstractive summarization. The three
convolutional models show comparable proportions of novel $n$-grams,
while RNN-based models show greater variance with \textsc{Seq2Seq}
generating the highest proportion of novel
$n$-grams. \textsc{PtGen+Covg} performs the least rewriting, followed
by \textsc{PtGen}. Interestingly, \textsc{PtGen} trained on XSum only
copies 4\% of 4-grams from the source document, 10\% of trigrams, 27\%
of bigrams, and 73\% of unigrams. This is in sharp contrast to
\textsc{PtGen} trained on CNN/DailyMail which copies more than 85\% of
4-grams in the source document, 90\% of trigrams, 95\% of bigrams, and
99\% of unigrams \cite{see-acl17}.  We should point out that the
summaries being evaluated have on average comparable lengths:
summaries generated by \textsc{Seq2Seq}, \textsc{PtGen}, and
\textsc{PtGen+Covg} contain 23.02, 22.86, and 22.52 words,
respectively; those generated by \textsc{ConvS2S},
\textsc{ConvS2S+Copy}, and \textsc{T-ConvS2S} have 20.07, 19.82 and
20.22 words, respectively, while \textsc{gold} summaries are the
longest with~23.26 words.




\begin{table}[t]
  \center{ 
    \begin{tabular}{lrc}
      \thickhline
      Models  & Score & QA\\ \hline
      \textsc{ext-oracle} &  -0.121 & 15.70\\
      \textsc{PtGen} & -0.218& 21.40 \\
      \textsc{ConvS2S}  &
      -0.130 & 30.90 \\
      \textsc{T-ConvS2S} & {0.037} & {46.05}\\ \hline
      \textsc{gold} & 0.431 & 97.23\\
      \thickhline
    \end{tabular}}
  \caption{System ranking according to human judgments  and QA-based
    evaluation for the XSum dataset.\label{tab:heval-pref-xsum}}
\end{table}

\paragraph{Human Evaluation}

Recall that system generated summaries were evaluated in two studies
one aimed at eliciting judgments of summary quality and the other
following a question-answering paradigm. In both studies, participants
were asked to evaluate summaries produced from the \textsc{ext-oracle}
baseline, \textsc{PtGen}, the best performing RNN-based system
according to ROUGE (see Table~\ref{tab:rouge-xsum}), \textsc{ConvS2S},
our topic-aware model \textsc{T-ConvS2S}, and the human-authored gold
summary (\textsc{gold}). We did not include summaries from the
\textsc{lead} or \textsc{ConvS2S+Copy} as they were significantly
inferior to other models. Table~\ref{tab:heval-pref-xsum} presents our
results.

Perhaps unsurprisingly human-authored summaries were considered best,
whereas, \textsc{T-ConvS2S} was ranked 2nd followed by
\textsc{ext-oracle} and \textsc{ConvS2S}. \textsc{PtGen} was ranked
worst with the lowest score of $-0.218$. We carried out pairwise
comparisons between all models to assess whether system differences
are statistically significant. \textsc{gold} is significantly
different from all other systems and \textsc{T-ConvS2S} is
significantly different from \textsc{ConvS2S} and \textsc{PtGen}
(using a one-way ANOVA with posthoc Tukey HSD tests; $p < 0.01$). All
other differences are not statistically significant.

The rightmost column in Table~\ref{tab:heval-pref-xsum} shows the
results of the QA evaluation. Based on the summaries generated by
\mbox{\textsc{T-ConvS2S}}, participants can answer $46.05\%$ of the
questions correctly. Summaries generated by \textsc{ConvS2S},
\textsc{PtGen}, and \textsc{ext-oracle} provide answers to $30.90\%$,
$21.40\%$, and $15.70\%$ of the questions, respectively. Pairwise
differences between all systems are statistically significant ($p <
0.01$) with the exception of \textsc{PtGen} and \textsc{ext-oracle}.
\textsc{ext-oracle} performs poorly on both QA and rating
evaluations. The examples in Figure~\ref{fig:erroranalysis-xsum}
indicate that \textsc{ext-oracle} is often misled by selecting a
sentence with the highest ROUGE (against the gold summary), but ROUGE
itself does not ensure that the summary retains the most important
information from the document.  The QA evaluation further emphasizes
that in order for the summary to be felicitous, information needs to
be embedded in the appropriate context. For example, \textsc{ConvS2S}
and \textsc{PtGen} will fail to answer the question ``Who has
resigned?''  (see Figure~\ref{fig:erroranalysis-xsum} second block)
despite containing the correct answer ``Dick Advocaat'' due to the
wrong context. \mbox{\textsc{T-ConvS2S}} is able to extract important
entities from the document with the right theme.

\subsection{Results on the Newsroom-Abs Dataset}
\label{sec:results-newsroom}

We next examine whether our approach extends to other datasets with
similar characteristics. Specifically, we examine the performance of
the proposed model and related models on the abstractive portion of
the Newsroom dataset (Newsroom-Abs; \citeR{newsroom-naacl18}).

\begin{table}[t]
  \begin{center}{
  \begin{tabular}{ l | c c c } 
    \thickhline
    Models & R1 & R2 & RL \\ \thickhline 

    \textsc{random} & 13.02 & 1.51 & 10.46 \\
    \textsc{lead} \cite{newsroom-naacl18} & 13.76 & 2.42 & 11.29 \\
    \textsc{lead} & 15.44 & 2.72 & 12.32 \\
    \textsc{ext-oracle} & {29.85} & {7.82} & {24.86}\\ 
    \hline\hline
    \textsc{PtGen} \cite{newsroom-naacl18} &  14.71 & 2.27 & 11.48 \\
    \textsc{Seq2Seq} & 15.23 & 4.21 & 12.88 \\
    \textsc{PtGen} & 17.61 & 5.15 & 14.73 \\
    \textsc{PtGen+Covg} & 16.13 & 4.33 & 13.47 \\ \hline\hline
    \textsc{ConvS2S} & 16.77 & 5.57 & 14.54\\
    \textsc{ConvS2S+Copy} & 16.31 & 5.34 & 14.24 \\
    \textsc{T-ConvS2S} & 16.97 & 5.56 & 14.70\\ \thickhline 

  \end{tabular}}
  \end{center}
  \caption{Results on  NewsRoom-Abs test set. We report ROUGE-1 (R1), ROUGE-2 (R2),  
    and ROUGE-L (RL) F$_1$ scores. Extractive systems are in the upper block, RNN-based abstractive systems 
    are in the second block, and our convolutional abstractive systems are in the third block. 
    \label{tab:rouge-newsroom}}
\end{table}

\paragraph{Automatic Evaluation} 

Table~\ref{tab:rouge-newsroom} summarizes the results of our
ROUGE-based evaluation. In addition to earlier discussed systems (see
Section~\ref{subsec:comparisons}), we also include
\citeS{newsroom-naacl18} versions of \textsc{lead} and \textsc{PtGen}.
Our \textsc{lead} selects the first 2~sentences to form the summary
compared to the \textsc{lead} reported in \citeA{newsroom-naacl18}
which selects the first 3~sentences. Our \textsc{lead} is more in line
with the average number of sentences observed in the reference
summaries which is~1.25 (see Table~\ref{table:bbc-size-comparison-2}),
and as result obtains better performance. We also found that
\textsc{PtGen} \cite{newsroom-naacl18} was trained on the whole
Newsroom dataset. To make a fair comparison, we report results with
models trained on the NewsRoom-Abs portion of the dataset only. The
discrepancy in the results between our \textsc{PtGen} and the
\textsc{PtGen} model reported in \newcite{newsroom-naacl18} can be
explained by the usage of different training sets.

Our convolutional models, \textsc{ConvS2S}, \textsc{ConvS2S+Copy}, and
\textsc{T-ConvS2S}, significantly outperform the \textsc{lead}
baselines.\footnote{Again, we use the \texttt{pyrouge} script to
  estimate statistical significance.} \textsc{ConvS2S} significantly
outperforms \textsc{Seq2Seq}, its \mbox{RNN} counterpart but lags
behind \textsc{PtGen}.  \textsc{T-ConvS2S} performs competitively
against \textsc{PtGen} on R1 and RL scores and better on R2 (5.56 vs
5.15). The superior performance of \textsc{T-ConvS2S} over
\textsc{ConvS2S} confirms our hypothesis that \textsc{T-ConvS2S}
enhanced with topic information is better at identifying pertinent
content and generating informative summaries. The worse performance of
\textsc{ConvS2S+Copy} against \textsc{ConvS2S} further supports our
claim that the multi-hop attention mechanism already in place in
\textsc{ConvS2S} is very effective at resolving UNKs simply by copying
the most attended words from the source. However, this is not case
with the \mbox{RNN-based} models \cite{see-acl17}. As previously
discussed, \textsc{Seq2Seq} is prone to generating UNKs (see the top
block in Figure~\ref{fig:erroranalysis-newsroom}), while
\textsc{PtGen} corrects for this with the copy mechanism. The
summaries generated by \textsc{Seq2Seq} have a total of~11,418 UNK
words, whil \textsc{PtGen} only generates 4,467~UNK words on the
Newsroom-Abs test set.\footnote{Surprisingly, none of the
  \mbox{RNN-based} abstractive models generates UNK words on
  XSum. We believe this is due to the smaller vocabulary size (81,092 for
  XSum vs. 157,939 for Newsroom-Abs;
  Table~\ref{table:bbc-size-comparison-2}).}


Interestingly, all abstractive summaries fall behind the extractive
oracle (\textsc{ext-oracle}) on this dataset. In contrast, most
abstractive models (\textsc{PtGen}, \textsc{ConvS2S},
\textsc{ConvS2S+Copy} and \textsc{T-ConvS2S}) were able to outperform
\textsc{ext-oracle} on Xsum.  This suggests that Newsroom-Abs still
has some bias towards extractive methods and improved sentence
selection would bring further performance gains for extractive
approaches on this dataset. Models trained on XSum are better at
generating good quality abstracts, e.g.,~\textsc{T-ConvS2S} achieves
ROUGE scores of 31.89/11.54/25.75 compared to 16.97/5.56/14.70 on
Newsroom-Abs. There are two reasons for this. Firstly, Newsroom has a
great variety of summarization styles due to its collection from
multiple news outlets; it is hard for abstractive methods to
effectively model this. Secondly, XSum reference summaries are more
representative of document content, whereas, Newsroom-Abs summaries
are more indicative (see Section~\ref{subsec:contentindicative} for
examples of reference summaries from XSum and Newsroom). It is
probably harder for abstractive models to generate indicative
summaries describing the source text rather directly presenting its
content.

\begin{figure}[t!]
  \center{\scriptsize 
    \begin{tabular}{l p{9.4cm} l}
      \thickhline 

      
      \textsc{lead} & Recently, I had to clear out all our kitchen cabinets to prep for a renovation. This taught me a few things: & [6.1, 0.0, 5.1]\\ 
      \textsc{ext-oracle} & Once I started trying to avoid chemical cleaners, I think my stash doubled. Here's how to make them: you can use either vinegar or lemon juice in this recipe. & [14.3, 0.0, 10.4]\\
      \textsc{Seq2Seq} & A new recipe for \textcolor{red}{[UNK] [UNK]} is a recipe for \textcolor{red}{[UNK] [UNK]}, \textcolor{red}{[UNK]}, \textcolor{red}{[UNK]}, \textcolor{red}{[UNK]}, \textcolor{red}{[UNK]}, \textcolor{red}{[UNK]}, \textcolor{red}{[UNK]} and \textcolor{red}{[UNK]}. & [10.0, 0.0, 9.8] \\ 
      \textsc{PtGen} & DIY recipes can help you find a way to enjoy the perfect lemon. Here's how to get it. & [13.3, 0.0, 11.8]\\
      \textsc{PtGen+Covg} & DIY simple recipes from your kitchen, and your kitchen cabinets are n't just a few things to do. & [6.9, 0.0, 6.2]\\
      \textsc{ConvS2S} & From DIY furniture to DIY kitchen products, here's how to get \textcolor{red}{rid} of your kitchen. & [7.7, 0.0, 7.4]\\
      \textsc{ConvS2S+Copy} & How to get \textcolor{red}{rid} of your kitchen. & [10.5, 0.0, 10.0]\\
      \textsc{T-ConvS2S} & \textcolor{red}{DIY} \textcolor{midnightblue}{DIY products} for \textcolor{midnightblue}{cleaning} up your kitchen. & [10.5, 0.0, 10.0] \\
      \textsc{gold} & \multicolumn{2}{l}{These effective \textcolor{midnightblue}{natural cleaners} can be made at reasonable costs }\\
      Questions &\multicolumn{2}{l}{(1) {What \textcolor{midnightblue}{type of cleaners} are being discussed in this article?} (\textcolor{midnightblue}{Natural cleaners})}\\
      
      \hline 
      \textsc{lead} & During a shaky performance in the witness box, \textcolor{midnightblue}{Pistorius} changed his story and his defence from previous statements read at his bail hearing and the start of his trial; adding the last words he exchanged with \textcolor{midnightblue}{Steenkamp} and that he heard the lavatory door slam, and first saying he fired at an intruder, then that he fired by mistake. \textcolor{midnightblue}{Oscar Pistorius} sits in the dock during his trial in Pretoria (Reuters) & [26.3, 4.8, 17.7] \\
      \textsc{ext-oracle} & On his account of challenging an intruder, \textcolor{midnightblue}{Pistorius} admitted he didn't check the sound he heard with his \textcolor{midnightblue}{girlfriend} lying awake next to him. Witness Dr Johan Stipp told the court he saw a light on in Pistorius' bathroom when he heard shots fired; the athlete said he was ``too scared'' to switch the lights on until after the shooting. A psychologist said his reaction might seem ``extraordinary'' for an able-bodied person, but could be explained by his disability. & [33.3, 5.6, 19.2] \\

      \textsc{Seq2Seq} & \textcolor{midnightblue}{Oscar Pistorius}'s decision to \textcolor{red}{remove} his \textcolor{midnightblue}{girlfriend Reeva Steenkamp} in the face of the \textcolor{midnightblue}{murder of Reeva Steenkamp} has been charged with murder.  & [31.2, 13.7, 20.1] \\
      \textsc{PtGen} & Five couples who have been killed in a shooting at a gun in Pretoria after the shooting death of the gun. & [16.4, 0.0, 9.9] \\
      \textsc{PtGen+Covg} & Five couples have been arrested in the dock trial of an intruder in the dock on the dock during a dock trial. & [23.3, 2.9, 12.3] \\
      
      \textsc{ConvS2S} & A day in the life of armed America. & [11.3, 0.0, 7.0] \\ 
      \textsc{ConvS2S+Copy} & \textcolor{midnightblue}{Oscar Pistorius}, \textcolor{red}{who was shot} in the shooting of \textcolor{midnightblue}{Reeva Steenkamp}, has been accused of killing his \textcolor{midnightblue}{girlfriend}, \textcolor{midnightblue}{Reeva Steenkamp}. & [34.9, 11.1, 22.5] \\
      \textsc{T-ConvS2S} & A day after the shooting of \textcolor{midnightblue}{Reeva Steenkamp}, \textcolor{midnightblue}{Oscar Pistorius} has been accused of murdering his \textcolor{midnightblue}{girlfriend Reeva Steenkamp} in the shooting of Reeva Steenkamp. & [34.9, 14.1, 22.5] \\
      \textsc{gold} & \multicolumn{2}{p{12cm}}{\textcolor{midnightblue}{Oscar Pistorius} is on trial for the \textcolor{midnightblue}{premeditated murder} of his \textcolor{midnightblue}{girlfriend Reeva Steenkamp}. The state say \textcolor{midnightblue}{he killed her deliberately after an argument}; the athlete says he believed she was an intruder. The following are 10 reasons why the judge might find him guilty, or innocent} \\
      Questions & \multicolumn{2}{p{12cm}}{Who is on trial for the \textcolor{midnightblue}{premeditated murder} of his girlfriend? (\textcolor{midnightblue}{Oscar Pistorius})} \\
      & \multicolumn{2}{p{12cm}}{Who was his \textcolor{midnightblue}{girlfriend}? (\textcolor{midnightblue}{Reeva Steenkamp})} \\
      & \multicolumn{2}{p{12cm}}{What is the state's point of view? (\textcolor{midnightblue}{he killed her deliberately after an argument})} \\
      \thickhline

    \end{tabular}     
  }
  \vspace{-0.4cm}
  \caption{\small Example output summaries on the Newsroom Abstractive
    test set with [ROUGE-1, ROUGE-2 and ROUGE-L] scores, gold standard
    reference, and corresponding questions. Words highlighted in blue
    are either the right answer or constitute appropriate context for
    inferring it; words in red lead to the wrong
    answer. }\label{fig:erroranalysis-newsroom}
\end{figure}
 
\begin{table}[t]
  \center{ 
    \begin{tabular}{lrc}
      \thickhline
      Models  & Score & QA\\ \hline
      \textsc{ext-oracle} &  {0.473} & {41.31}\\
      \textsc{PtGen} & -0.047 & 20.98 \\
      \textsc{ConvS2S}  & -0.397 & 11.97 \\
      \textsc{T-ConvS2S} & -0.160 & 23.28\\ \hline
      \textsc{gold} & 0.130 & 90.98\\
      \thickhline
    \end{tabular}}
  \caption{System ranking according to human judgments  and QA-based
    evaluation for the Newsroom-Abs dataset.\label{tab:heval-pref-newsroom}}
\end{table}

\paragraph{Human Evaluation}

For both evaluation protocols, participants were asked to assess
summaries produced from the \textsc{ext-oracle} baseline,
\textsc{PtGen}, \textsc{ConvS2S}, our topic-aware model
\textsc{T-ConvS2S}, and the human-authored gold summary
(\textsc{gold}). We did not include summaries from the \textsc{lead},
or \textsc{ConvS2S+Copy} as they were significantly inferior to other
models. Table~\ref{tab:heval-pref-newsroom} presents our results.

To our surprise, \textsc{ext-oracle} summaries were considered best,
whereas the human-authored summaries were ranked 2nd followed by
\textsc{PtGen} and \textsc{T-ConvS2S}. \textsc{ConvS2S} was ranked
worst with the lowest score of $-0.397$. In line with our findings in
Section~\ref{subsec:contentindicative}, participants found
\textsc{ext-oracle} summaries to be more informative than
human-authored summaries which are often indicative in nature.  We
carried out pairwise comparisons between all models to assess whether
system differences are statistically significant. The difference
between \textsc{T-ConvS2S} and \textsc{PtGen} is not statistically
significant (using a one-way ANOVA with posthoc Tukey HSD tests; $p <
0.01$), while all other differences are.


The rightmost column in Table~\ref{tab:heval-pref-newsroom} shows the
results of the QA evaluation. Based on the oracle extracts,
participants can answer $41.31\%$ of the questions
correctly. Summaries generated by \textsc{T-ConvS2S}, \textsc{PtGen},
and \textsc{ConvS2S} provide answers to $23.28\%$, $20.98\%$, and
$11.97\%$ of the questions, respectively. \textsc{ext-oracle} performs
best on both QA and judgment elicitation evaluations. However, we
should point out that \textsc{ext-oracle} has the advantage of
selecting the best set of sentences (as determined by ROUGE) without
any length constraints. Consequently, summaries generated by
\textsc{ext-oracle} tend to be longer with~38.84 words on average (see
Figure~\ref{fig:erroranalysis-newsroom} second block). Summaries
generated by \textsc{PtGen}, \textsc{ConvS2S}, and \textsc{T-ConvS2S}
contain 23.72, 19.41 and 20.66 words, respectively, while
\textsc{gold} summaries contain 23.26 words. Perhaps unsurprisingly
\textsc{T-ConvS2S}, which was slightly lagging behind \textsc{PtGen}
on R1 and RL scores, performs better than both \textsc{PtGen} and
\textsc{ConvS2S} in terms of correctly answering questions. The QA
evaluation shows that \textsc{T-ConvS2S} is able to generate
informative summaries with pertinent information embedded in the
appropriate context. Pairwise differences between systems are all
statistically significant ($p < 0.01$) with the exception of
\textsc{T-ConvS2S} and \textsc{PtGen}, and, \textsc{PtGen} and
\textsc{ConvS2S}.




\begin{figure}[t!]
  \center{\footnotesize 
    \begin{tabular}{l p{12.5cm} r}
      \thickhline 
      \multicolumn{2}{c}{Reference Summaries} & Score \\ \thickhline 
      \multirow{26}{*}{\rotatebox[origin=c]{90}{XSum}} & Match reports from Saturday's Scottish Premiership and Championship games.  &  1.67 \\ 
      &  The case for raising US interest rates has ``strengthened'', the head of the Federal Reserve has said.  &  2.00 \\ 
      &  Internet giant Amazon's owner Jeff Bezos has made an amazing underwater discovery.  &  2.00 \\ 
      &  Deputy First Minister Martin Mcguinness pulled out of a trip to China this week due to medical advice, it has been revealed.  &  2.00 \\ 
      &  The Conservatives have accused Labour and Plaid Cymru of being involved in an ``unedifying squabble'' over who to support if there is a hung parliament.  &  2.00 \\ 
      &  Police have released CCTV images of a man they want to speak to following a racist assault on a Glasgow bus.  &  2.33 \\ 
      &  A yellow `be aware' weather warning was re-introduced in parts of Wales on Sunday.  &  2.33 \\ 
      &  Ross County's Jim Mcintyre is the only Premiership boss in the running for PFA Scotland's manager of the year award.  &  2.33 \\ 
      &  A council plans to employ its own staff to help young people with mental health problems.  &  2.33 \\ 
      &  UK-born Oliver Hart and Bengt Holmstrom of Finland have won the Nobel Economics prize for work on Contract Theory.  &  2.33 \\ 
      &  Leinster resisted a spirited Cardiff Blues revival to strengthen their position at the top of the Pro12 table.  &  2.33 \\ 
      &  DNA analysis of a 45,000-year-old human has helped scientists pinpoint when our ancestors interbred with Neanderthals.  &  2.33 \\ 
      &  Tributes have been paid to a former Glamorgan cricketer who was found dead at his Swansea flat on friday.  &  2.33 \\ 
      &  A new visitor centre at the battle of Britain memorial -- designed in the shape of a spitfire wing -- is beginning to take shape.  &  2.67 \\ 
      &  The first of the missing Nigerian schoolgirls to be rescued since her capture two years ago has had an emotional reunion with her mother.  &  2.67 \\ 

      \hline\hline

      \multirow{20}{*}{\rotatebox[origin=c]{90}{Newsroom-Abs}} & The pair costarred in a Star Wars-themed sketch for an unaired TV show in 2008  &  1.33 \\ 
      &  Things were beginning to get ugly in the Democratic race  &  1.33 \\ 
      &  Installations will take place in 100 stores.  & 1.33 \\ 
      &  Sean Penn, actor, director, hothead, misanthrope. So what's he still doing in haiti?  & 1.33 \\ 
      &  Collection of all usatoday.com coverage of the Exorcist, including articles, videos, photos, and quotes.  & 1.33 \\ 
      &  Bain wasn't just about private equity.  & 1.33 \\ 
      &  An expert's advice for college students and parents  & 1.33 \\ 
      &  The 62nd instalment in a weekly series that debunks the web's hoaxes, rumors and exaggerations.  & 1.67 \\ 
      &  The ability to cope with risk and work insane hours comes in handy. Take it from these former financiers.  & 1.67 \\ 
      &  The FED needs to stop this cat-and-mouse game and just say it isn't raising rates anytime soon, says Carol Roth.  & 2.00 \\ 
      &  The government hopes everyone has exchanged their old notes by now.  &  2.00 \\ 
      &  Washington Post food critic Tom Sietsema takes a fresh look at 13 restaurants he has previously reviewed. How do they fare the second time around?  & 2.00 \\ 
      &  Biden's son had mild stroke -- the Oval : tracking the Obama presidency  & 2.00 \\ 
      &  The cost of home shopping could soar as top delivery company ups its charges  & 2.00 \\ 
      &  Probiotic bacteria can affect behavior.  &  2.33 \\ 
      
      \thickhline

    \end{tabular}     
  }
  \vspace{-0.4cm}
  \caption{Summaries ranked from least to most
 informative from the XSum and Newsroom-Abs test
    sets. We also show their informativeness scores (higher is better).
  }\label{fig:summaryanalysis-xsum-newsroom}
\end{figure}

\subsection{Informative vs. Indicative Summaries}
\label{subsec:contentindicative}

Our experiments have so far revealed differences in the nature of XSum
and Newsroom summaries.  XSum summaries are often informative of the
document content while Newsroom summaries are indicative, i.e.,~they
describe the source text rather than directly presenting the
information it contains. In this section, we provide further empirical
support for this claim.

\begin{table}[t!]
  \center{
    \begin{tabular}{l | c c c | r}
      \thickhline
      Dataset  & Informative &  Part. Informative & Uninformative & Score \\ \hline
      XSum &  {68.00} & 26.00 & \hspace*{.2cm}6.00 & {2.62}\\
      Newsroom-Abs & 48.67 & 33.33 & 18.00 & 2.30 \\
      \thickhline
    \end{tabular}}
  \caption{XSum and Newsroom-Abs  summaries and their
    informativeness. The middle column proportionately shows the number 
    times a summary  was judged ``Informative'', 
    ``Partially Informative'', or ``Uninformative.'' The last column 
    shows the informativeness score for each dataset 
    (higher is  better). \label{tab:heval-content-indicative}}
\end{table}

We conducted a human evaluation where participants were asked to rate
reference summaries from both datasets. Specifically, they were
presented with a document and its gold summary and asked to decide
whether it was informative (i.e.,~it relayed pertinent content from
the document), partially informative, or uninformative. The study was
conducted on Amazon Mechanical Turk with the same 100~test documents
used of our judgment elicitation and QA studies on XSum and
Newsroom-Abs.  We collected judgments from three different
participants for each document. The order of documents and systems
were randomized.

\pgfplotstableread[row sep=\\,col sep=&]{
  ngrams & XSum & Newsroom\\ 
  1&99.3292028213&98.9521464036 \\
  2&86.8296376763&81.1293220034  \\
  3&57.0633335962&43.604015835  \\
  4&31.1966928901&19.5852985453  \\
  5&16.6180183565&10.6742752479 \\
  6&9.03651434233&07.6621975706  \\
  7&5.1470701018&06.49293422207  \\
  8&3.1929845506&05.85944596059  \\
  9&2.14187849392&05.46259876106  \\
  10&1.51341790002&05.13326383827  \\
  11&1.15809067943&04.78702623548  \\
  12&0.942684495283&04.4469876071  \\
  13&0.798462451952&04.25236900084  \\
  14&0.687767322498&04.16313445599  \\
  15&0.595755126704& 04.07797612167 \\
}\ngramsseen

\begin{figure}[t!]
  \center{
    \begin{tikzpicture}
      \begin{axis}[
        title={},
        xlabel={$n$-grams (in test summaries)},
        ylabel={Percentage (in training summaries)},
        xmin=0, xmax=15,
        ymin=0, ymax=100,
        xtick={1,2,3,4,5,6,7,8,9,10,11,12,13,14,15},
        ytick={0, 10, 20, 30, 40, 50, 60, 70, 80, 90, 100},
        legend pos=north east,
        xmajorgrids=true,
        ymajorgrids=true,
        grid style=dashed,
        ]
        \addplot [midnightblue] table[x=ngrams,y=XSum]{\ngramsseen};
        \addplot [purple] table[x=ngrams,y=Newsroom]{\ngramsseen};
        \legend{XSum,Newsroom-Abs}
      \end{axis}
    \end{tikzpicture}
  }
  \caption{Percentage of $n$-grams in test summaries seen in training
    summaries.\label{fig:seenngrams}}
\end{figure}
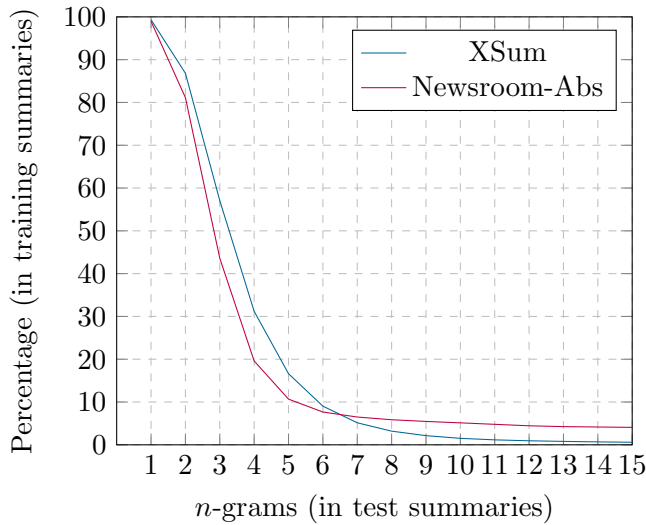

Table~\ref{tab:heval-content-indicative} present the results of this
study.  We measure the informativeness of each dataset as the average
score assigned by crowdworkers across summaries; a summary receives a
score of 3 if it is deemed informative, 2 if it is partially
informative, and 1 if it is uninformative. Therefore, the
informativeness score for a dataset as a whole may vary from 1 to 3,
with 1 being least informative and 3 being most informative. We also
report the proportion of times a dataset was considered informative,
partially informative, and uninformative.  XSum reference summaries
were mostly considered informative (68\%), around a quarter of them
(26\%) were deemed partially informative, and only 6\% were found
uninformative. In comparison, less than half (48.67\%) of Newsroom-Abs
summaries were found informative, the remaining being either partially
informative or uninformative. XSum achieved an informativeness score
of~2.62 compared to~2.30 for the Newsroom dataset. We carried out
pairwise comparisons to assess whether informativeness differences are
statistically significant. We found that XSum is significantly more
informative than Newsroom (using a one-way ANOVA with posthoc Tukey
HSD tests; $p < 0.01$). Figure~\ref{fig:summaryanalysis-xsum-newsroom}
shows the 15 best summaries from each dataset ranked from least to
most informative.


\pgfplotstableread[row sep=\\,col sep=&]{
  ngrams & XSum & Newsroom\\ 
  1 & 0.0164456192539 & 0.0195532801163 \\
  2 & 0.196666771225 & 0.24921002522 \\
  3 & 0.519871037028 & 0.630312440491 \\
  4 & 0.755612190807 & 0.836118212928 \\
  5 & 0.876251643306 & 0.906063631767 \\
  6 & 0.935171465359 & 0.928924453492 \\
  7 & 0.964009190534 & 0.938260481132 \\
  8 & 0.978307991915 & 0.943441887403 \\
  9 & 0.985744702177 & 0.946946679309 \\
  10 & 0.989852801032 & 0.949985399146 \\
  11 & 0.992171590603 & 0.953216795268 \\
  12 & 0.993642666146 & 0.956448038227 \\
  13 & 0.994638371089 & 0.958160818236 \\
  14 & 0.995317516489 & 0.959078884389 \\
  15 & 0.995831755167 & 0.960003403203 \\
}\typetoken

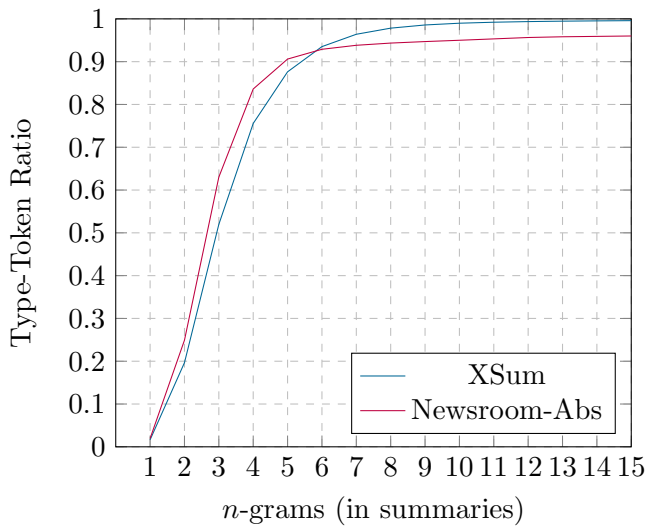
\begin{figure}[t!]
  \center{
    \begin{tikzpicture}
      \begin{axis}[
        title={},
        xlabel={$n$-grams (in summaries)},
        ylabel={Type-Token Ratio},
        xmin=0, xmax=15,
        ymin=0, ymax=1,
        xtick={1,2,3,4,5,6,7,8,9,10,11,12,13,14,15},
        ytick={0, 0.10, 0.20, 0.30, 0.40, 0.50, 0.60, 0.70, 0.80, 0.90, 1.00},
        legend pos=south east,
        xmajorgrids=true,
        ymajorgrids=true,
        grid style=dashed,
        ]
        \addplot [midnightblue] table[x=ngrams,y=XSum]{\typetoken};
        \addplot [purple] table[x=ngrams,y=Newsroom]{\typetoken};
        \legend{XSum,Newsroom-Abs}
      \end{axis}
    \end{tikzpicture}
  }
  \caption{Type-Token Ratio for summary $n$-grams in the
    entire dataset.\label{fig:typetokenratio}}
\end{figure}


Figure~\ref{fig:seenngrams} shows the percentage of $n$-grams in test
summaries which have been already seen in training summaries. For both
datasets, as the size of $n$-grams increases, their chance of having
been seen in the training summaries decreases rapidly. For XSum, this
percentage drops to almost 0\% for any $n$-grams larger than
size~10. Interestingly, this is not the case for Newsroom-Abs, where
more than 4\% of $n$-grams (sizes 10 to 15) in test summaries have
been already seen in training summaries. This result suggests that
Newsroom-Abs summaries are somewhat formulaic displaying a certain
degree of repetition that goes beyond simple phrases.  As an example
consider the summary from
Figure~\ref{fig:summaryanalysis-xsum-newsroom} ``Collection of all
usatoday.com coverage of the Exorcist, including articles, videos,
photos, and quotes.'' which is rather generic and would apply to any
movie, not just the Exorcist. In fact two crowdworkers labeled this
summary as uninformative and one as partially informative
(informativeness score is~1.33). Figure~\ref{fig:typetokenratio} shows
the type-token ratio of different $n$-grams in gold summaries as a
measure of how often constructions are being reused; a higher
type-token ratio represents larger variation in terms of
$n$-grams. XSum summaries exhibit more variation for $n$-grams of size
larger than 5 corraborating our claim that Newsroom-Abs summaries are
more repetitive.



\section{Conclusions}

In this paper we introduced the task of ``extreme summarization''
together with a large-scale dataset which pushes the boundaries of
abstractive methods. We further proposed a novel ``topic-aware''
fully-convolutional deep learning model which is well-suited to
extreme summarization. And designed a question-answering paradigm to
assess the degree to which abstractive models retain key information
from the document.  Experimental evaluation revealed that models which
have abstractive capabilities do better on this task and that
high-level document knowledge in terms of topics and long-range
dependencies is critical for recognizing pertinent content and
generating informative summaries. Finally, experimental results
support our claim that extreme summarization is a good testbed for
abstractive summarization; the task, as operationalized via our
dataset, encourages models to create informative summaries which
promote novel constructions and are less skewed towards extractive
mechanisms.


Extreme summarization revisits interesting problems in abstractive
summarization with the relatively simpler objective of generating
single sentence summaries rather than multi-line summaries
\cite{N18-2097,wiki:iclr:18,yasunaga.aaai19.scisumm}. Models trained
for extreme summarization require document-level inference,
abstraction, and paraphrasing to generate summaries which are
informative and consistent with the input document. Throughout this
paper we have argued that our model is better suited for this task
than recurrent abstractive models due to its ability to foreground
pertinent content using topic vectors and model long-range
dependencies using a multi-layer convolutional architecture. In the
future, we would like to create more linguistically-aware encoders and
decoders incorporating co-reference and entity linking. It would also
be interesting to use contextualised word representations
\cite{peters-EtAl:2018:N18-1,bert,xlnet_arxiv19} to enhance modeling of long-range
dependencies within our model.

Beyond generating single sentences, we would like to adapt our method
to create multi-sentence summaries. For instance, this would allow us
to assess whether our model's ability to capture long-range
dependencies translates to more readable and coherent summaries.
Finally, our method might be relevant to summarizing texts from other
domains, e.g.,~generating multi-line abstracts of scientific articles
\cite{N18-2097,yasunaga.aaai19.scisumm}, creating Wikipedia pages in a
multi-document summarization setting \cite{wiki:iclr:18}, and
aggregating product or movie reviews
\cite{N16-1007,angelidis18summarizing}.


\paragraph{Acknowledgments} We thank the reviewers for their enthusiastic feedback. 
We gratefully acknowledge the support of
the European Research Council (Lapata; award number 681760), the
European Union under the Horizon 2020 SUMMA project (Narayan, Cohen;
grant agreement 688139) and Bloomberg (Cohen).



\vskip 0.2in
\bibliography{summarisation-improved}
\bibliographystyle{theapa}

\end{document}